\documentclass[sigconf]{acmart}
\AtBeginDocument{%
  }

\copyrightyear{2026}
\acmYear{2026}
\acmDOI{XXXXXXX.XXXXXXX}
\acmConference[Conference acronym 'XX]{Make sure to enter the correct
  conference title from your rights confirmation email}{June 03--05,
  2016}{Woodstock, NY}
\acmISBN{978-1-4503-XXXX-X/2018/06}
\usepackage{amsmath}
\usepackage{algorithm}
\usepackage{algorithmic}
\usepackage{multirow}
\usepackage{graphicx}
\usepackage{enumitem}
\PassOptionsToPackage{hyphens}{url}
\usepackage{hyperref}
\usepackage[capitalise]{cleveref}
\usepackage{float}
\usepackage{enumitem}

\newif\ifanonymous 
\anonymoustrue
\begin{document}
\newcommand{\codelink}{%
  \ifanonymous
    \url{anonymized-for-review}
  \else
    \url{https://github.com/JanMarcoRuizdeVargas/}
  \fi
}

\title{Counter-Dyna: Data-Efficient RL-Based HVAC Control using Counterfactual Building Models}



\author{Jan Marco Ruiz de Vargas}
\email{janmarco.ruiz@tum.de}
\orcid{0009-0001-4812-8666}
\affiliation{%
  \institution{Technical University of Munich}
  \city{Munich}
  \country{Germany}
}

\author{Fabian Raisch}
\orcid{0000-0003-1869-9801}
\affiliation{%
  \institution{Technical University of Applied Sciences Rosenheim}
  \city{Rosenheim}
  \country{Germany}
}
\affiliation{%
  \institution{Technical University of Munich}
  \city{Munich}
  \country{Germany}
}

\author{Zoltan Nagy}
\email{z.nagy@tue.nl}
\orcid{0000-0002-6014-3228}
\affiliation{%
  \institution{Eindhoven University of Technology}
  \city{Eindhoven}
  \country{Netherlands}
}

\author{Pierre Pinson}
\email{p.pinson@imperial.ac.uk}
\orcid{0000-0002-1480-0282}
\affiliation{%
  \institution{Imperial College London}
  \city{London}
  \country{United Kingdom}
}

\author{Christoph Goebel}
\email{christoph.goebel@tum.de}
\orcid{0000-0002-5756-6983}
\affiliation{%
  \institution{Technical University of Munich}
  \city{Munich}
  \country{Germany}
}

\renewcommand{\shortauthors}{Ruiz de Vargas et al.}

\begin{abstract}
Model-based reinforcement learning (MBRL) offers a promising approach for data-efficient energy management in buildings, combining the strengths of predictive modeling and reinforcement learning. While previous MBRL methods applied to HVAC control have reduced training data requirements, they still require several months of interaction with the building to learn a satisfactory control policy. A key reason is that existing surrogate models attempt to predict the entire state-space, including weather and electricity prices that are unaffected by control actions, or completely ignore these variables. Addressing these issues, we propose Counter-Dyna, a method that enhances the data-efficiency of Dyna, an MBRL method. We create data-efficient counterfactual surrogate models (CSM) by leveraging invariances in the state-space. Using a CSM in Dyna speeds up RL training measured in environment interaction data compared to previous results. In comparison with previous state-of-the-art that used 6-12 months of environment interactions, our method needs only 5 weeks. We evaluate our method in a large simulation study using the literature standard BOPTEST framework and proximal policy algorithm (PPO) as the RL algorithm. Our results show cost-saving potentials of 5.3\% to 17.0\% in a hypothetical deployment scenario. Our work is a significant step towards making real-world deployment of RL algorithms in HVAC control practically viable. 

\end{abstract}



\maketitle

\section{Introduction}
\label{sec:introduction}
Buildings account for about 30 \% of global CO2 emissions \cite{abergel2017towards}. In recent years, buildings are changing from passive consumers of energy to controllable grid assets, with great potential for reducing energy costs. Optimizing building energy use is becoming more challenging, due to the introduction of dynamic pricing and affordable residential photovoltaic and battery assets. Intelligent heating, ventilation, air conditioning and cooling (HVAC) control contributes to this by shifting heating or cooling energy use to low-cost periods, which are more pronounced due to an increasing share of intermittent renewable electricity generation \cite{vazquez2019reinforcement}. 

Reinforcement learning (RL) has emerged as a method for intelligently controlling buildings \cite{nagy2023tenquestionsrl}. RL can achieve performance close to a theoretical optimum of 96.54 \% w.r.t. a combination of cost, comfort and control slew \cite{fu2023good}. Unfortunately, achieving this performance requires the RL agent to interact with the building for years. To address this, model-based RL has been proposed \cite{gao2023comparative,saeed2024dyna}. Here, a surrogate model (like a digital twin) of the environment is fit with a data-driven method, which generates additional data for the RL agent, reducing the need for expensive and slow real interactions. 

While this previous work has shown that model-based RL (MBRL) is more sample efficient than model-free versions, a disconnect remains between academic state-of-the-art and the practical constraints of the building industry. The data efficiency of MBRL is still insufficient for rapid deployment, needing 6-12 months of building interaction \cite{gao2023comparative,saeed2024dyna}. 

Additionally, the simulation environments used, such as BOPTEST-gym \cite{arroyo2022openai}, allows episode resets to arbitrary points. This lets the control agents time-travel during training, which is not realistic. Furthermore, general model-based methods \cite{janner2019trust}, e.g. used in Gao et al. \cite{gao2023comparative}, forecast the entire state-space and rewards, which isn't necessary in this setting, because parts of the state-space are invariant to the control actions, as we will later show in \cref{sec:counterfactual_surrogate_model}. Including e.g. price and weather forecasts as prediction targets for the surrogate environment model amplifies the potential for model bias. This reduces the usefulness of the surrogate model. Other works like of Saeed et al. \cite{saeed2024dyna} don't make use of price or weather forecasts, which however is important information for the control agent to do flexible load shifting. 

We address these gaps by proposing an adaptation of the Dyna algorithm, which we call Counter-Dyna. Our contributions are:
\begin{itemize}[nosep]
    \item We rethink how surrogate modeling is done in Dyna-style reinforcement learning. Our counterfactual surrogate model (CSM) uses a causal lens on the state-space and only forecasts variables dependent on control actions for the CSM rollouts, while leaving invariant ones (like weather or price forecasts) unchanged.
    \item We impose a new and realistic training regime on the RL algorithms. Our training episodes happen in strict chronological order, compared to 'time-traveling' RL agents of previous works. This imitates a realistic deployment scenario. 
    \item We demonstrate the stability and data efficiency of Counter-Dyna compared to model-free methods in a large simulation study using PPO and SAC, two state-of-the-art RL algorithms. We analyze the accuracy of the CSM on counterfactual action sequences and investigate the sensitivity of Counter-Dyna to hyperparameters. 
\end{itemize}

The remainder of this paper is organized  as follows. In \cref{sec:related_work} we discuss the related work in more detail. \cref{sec:methodology} presents our methodology, in particular how we define the CSM, how we use it to run Dyna-style reinforcement learning. \cref{sec:evaluation} describes our experimental setup using BOPTEST and contains the results of our simulation studies, comparing Counter-Dyna with baseline controllers and model-free RL algorithms. We focus on the sample efficiency w.r.t. real environment interactions, cost and occupant discomfort. We discuss our results and suggest future research directions. \cref{sec:conclusion} summarizes our contributions and results. The nomenclature of our paper is summarized in \cref{tab:nomenclature}. 

\begin{table}[h!]
\centering
\caption{Nomenclature}
\label{tab:nomenclature}
\renewcommand{\arraystretch}{1.2}
\begin{tabular}{l p{7cm}}
\hline
\textbf{Symbol} & \textbf{Description} \\
\hline
CSM & Counterfactual Surrogate Model, used to generate model rollouts (=synthetic episodes) \\
HVAC & Heating, Ventilation, Air Conditioning \\
(MB)RL & (Model-based) Reinforcement Learning \\
MPC & Model Predictive Control \\
MLP & Multi-layer Perceptron \\
DDQN & Double deep Q-learning \\
PPO & Proximal Policy Optimization \\
SAC & Soft Actor-Critic \\
\hline
$a_t$ & Action at time step $t$ (real environment) \\
$\tilde{a}_l$ & Synthetic action at time step $l$ (taken in CSM) \\
$r_t$ & Reward at time step $t$ (real environment) \\
$\tilde{r}_l$ & Synthetic reward at time step $l$ \\
$s_t^{(n)}$ & State at time step $t$ within episode $n$ (from CSM) \\
$\tilde{s}_l$ & Synthetic state at time step $l$ (generated by CSM)\\
$z_t$ & Zone (building) temperature at time $t$ \\
$d_t$ & Forecasts \& disturbances at time $t$ (e.g., weather, price) \\
$D_{env}$ & Buffer of real environment episodes \\
$D_{model}$ & Buffer of CSM rollouts \\
$N$ & Total number of real environment episodes \\
$T$ & Length of a real episode \\
$K$ & Number of CSM rollouts per real episode \\
$L$ & Length of a CSM rollout (synthetic episode) \\
$m_\psi$ & Environment model parameterized by $\psi$ \\
$m_\beta$ & Reward model \\
\hline
\end{tabular}
\end{table}

\section{Related Work}
\label{sec:related_work}
In recent years, the research literature has proposed two main ways of replacing traditional, sub-optimal rule-based control: model predictive control (MPC) \cite{drgovna2020all} and deep reinforcement learning (DRL) \cite{wei2017deep}. Fu et al. \cite{fu2023good} show that, in principle, both MPC and DRL can achieve performance close to the theoretical optimum ($>$90\%, when optimizing for a combination of energy cost, thermal comfort and control slew rates). To test these advanced controllers, the building optimization testing framework (BOPTEST) \cite{blum2021building} has been developed, which we will also use to evaluate our method. BOPTEST is a standard way to evaluate HVAC control algorithms, modeling several buildings with high-fidelity Modelica simulations. It also comes with a gym interface \cite{arroyo2022openai}, which is suited for RL training. BOPTEST and BOPTEST-gym are open-source. 

\textbf{Model Predictive Control for HVAC Control.} MPC has demonstrated strong results and has a large body of literature supporting it \cite{drgovna2020all}. Unfortunately, MPC has problems for scalable (both large multi-zone buildings and many buildings), real-world deployment. For one, MPC relies on an accurate building model, which either requires expert knowledge and time to craft \cite{gao2019building} or historical, environment exploring data to fit a data-driven building model \cite{SMARRA20181252, raisch2025gentl}. Furthermore, depending on the resulting optimization problem, computing a solution might be costly and slow \cite{rao2009survey}. 

\textbf{Deep Reinforcement Learning for HVAC Control.} DRL offers an alternative, possibly purely data-driven method that can cheaply choose actions during inference. The training signal it uses, the reward, can incorporate almost any kind of information. This makes it very flexible and scalable. RL can be split into two categories: model-free and model-based. 
Model-free RL (MFRL) is attractive as it can learn purely from interacting with the environment. Wei et al. \cite{wei2017deep} use a model-free Deep Q-Network (DQN) \cite{mnih2013playing}, achieving significant energy cost reduction (19.1\%-71.2\% for different test cases). Similar results have been reported in numerous other studies. Most work has focused on the model-free approaches Q-learning \cite{wei2017deep} and actor-critic methods \cite{kathirgamanathan2021development}, according to several review papers \cite{han2019review,vazquez2019reinforcement,wang2020reinforcement,yu2021review,fu2022applications,al2024reinforcement}. 
However, model-free approaches face a significant barrier for deployment in real buildings: they require extensive interaction with the environment until they converge to stable operation. Hence, MFRL is slow, expensive and often not possible for an occupied building, as training could drive the system into unsafe states.
While safety guarantees can be added \cite{liu2022safe,zhang2022safe}, sample efficiency is still a problem. Model-free RL agents need years or decades of data to converge \cite{wang2023comparison,WANG2025124328}. This wastes time, energy and money. 

\begin{figure*}[t]
    \centering
    \begin{minipage}{0.54\textwidth}
        \centering
        \includegraphics[width=\textwidth, trim={0cm 0cm 0cm 0cm}, clip]{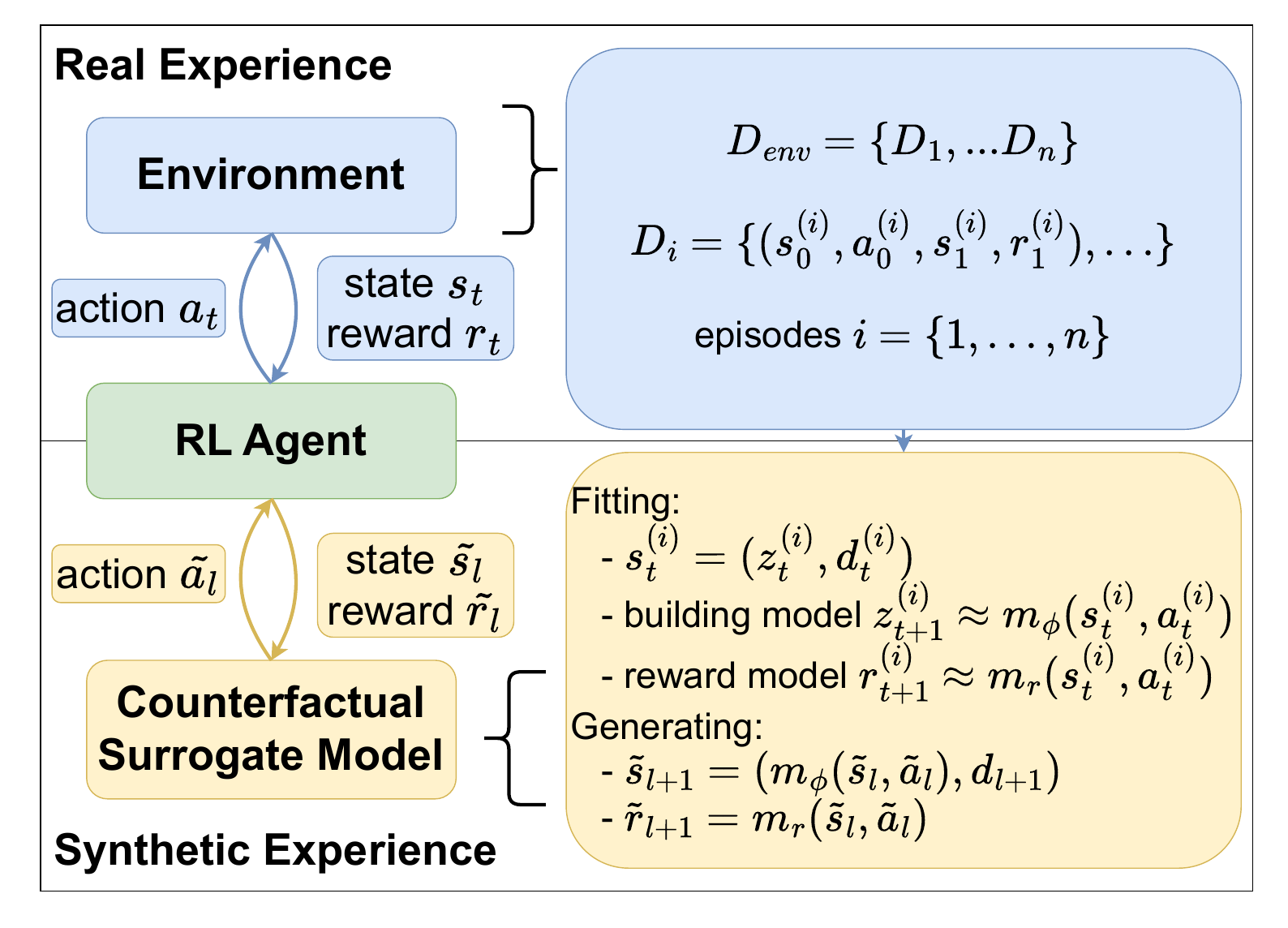}
    \end{minipage}
    \begin{minipage}{0.38\textwidth}
        \centering
        \includegraphics[width=\textwidth, trim={0cm 0cm 0cm 0cm}, clip]{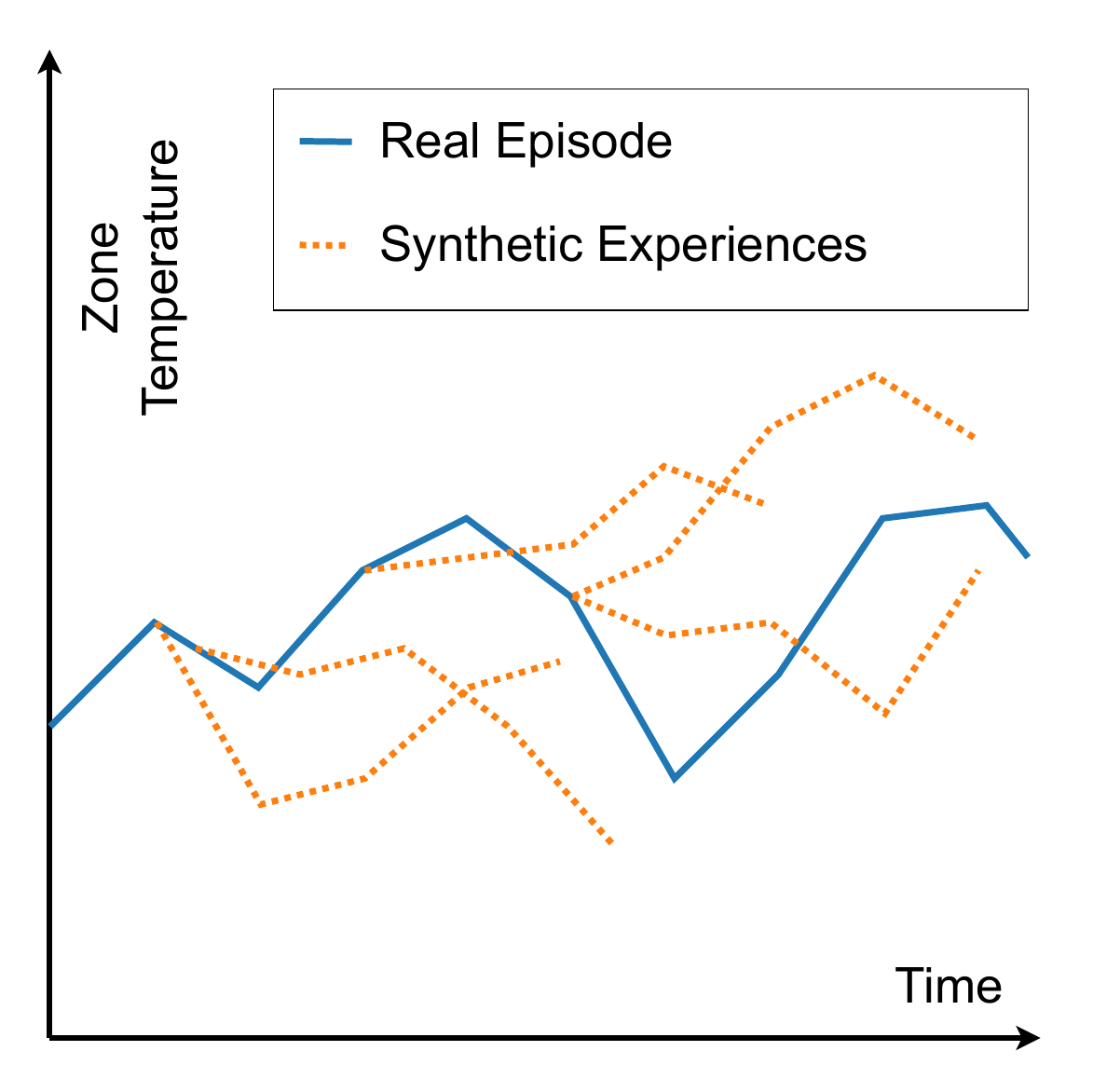}
    \end{minipage}

    \caption{\textbf{Left:} Flowchart illustrating the Dyna workflow. \textbf{Right:} Based on a point in the real episode (blue), the surrogate model allows the RL agent to simulate alternative trajectories/episodes (orange dotted).}
    \label{fig:dynambrl_visualized}
\end{figure*}

To address this, model-based RL (MBRL) has been proposed. Model-based RL doesn't have a uniform definition. In general, when a model of the environment is used to support RL fitting, the algorithm is called model-based. Several works follow a paradigm of fitting a model on historical data and then using the surrogate model to determine the policy with search algorithms, while improving the surrogate model simultaneously, using differential optimization or search \cite{chen2019gnurl,ding2020mb2c,chen2022mbrl,ding2025safe}. Another way is to first fit a surrogate model on historical data, and then use only the surrogate environment as the training environment for RL. 
\textbf{Dyna-style MBRL.} 
It is also possible to fit the building model and RL agent iteratively, which is called Dyna-style reinforcement learning \cite{sutton1991dyna}. We follow this approach. It allows the control agent to benefit from both real interaction and synthetic ones generated from the surrogate model, see \cref{fig:dynambrl_visualized,alg:counter_dyna}. Its policy can be a simple forward pass through a multi-layer perceptron (MLP), which is simpler than the aforementioned search or optimization-based methods. Additionally, it can integrate flexible and powerful model-free RL algorithms like DQN, SAC or PPO with a surrogate model, thus making the overall pipeline model-based. Dyna has been applied to HVAC control. 

\begin{algorithm}[htbp]
\caption{Counter-Dyna}
\label{alg:counter_dyna}
\begin{algorithmic}[1]
\STATE \textbf{Input:} initial policy $\pi_\theta:s_t\rightarrow a_t$, real environment dynamics $P(s_{t+1}|s_t,a_t)$, reward function $R(s_t,a_t,s_{t+1})$, initial counterfactual surrogate model $m: (s_t, a_t) \rightarrow (s_{t+1}, r_{t+1})$, initialized real experience buffer $D_{env} = \emptyset$ and synthetic experience buffer $D_{model} = \emptyset$. 
\FOR{$n \in \{1,\dots,N\}$ episodes of real environment interaction}

    \STATE get initial environment state $s_0^{(n)}$
    \FOR{$t \in \{0,\dots,T-1\}$ steps per real episode}
        \STATE sample action $a_{t}^{(n)} \sim \pi_\theta(a|s_t^{(n)})$
        \STATE real env. returns $s_{t+1}^{(n)} \sim P(s_{t+1}, r_{t+1}|s_t^{(n)}, a_t^{(n)})$
        \STATE get reward $r_{t+1}^{(n)} = R(s_t^{(n)}, a_t^{(n)},s_{t+1}^{(n)}) $
    \ENDFOR
    
    \STATE $D_n \gets \{(s_0^{(n)},a_0^{(n)},s_1^{(n)},r_1^{(n)}),\dots,(s_{T-1}^{(n)},a_{T-1}^{(n)},s_T^{(n)},r_T^{(n)})\} $
    \STATE add episode buffer to experience buffer $D_{env} \gets D_{env} \cup \{D_n\}$
    \STATE apply RL to $\pi_\psi$ on experiences $D_{env} \cup D_{model}$
    \STATE fit $m_\psi, m_\beta$ on $D_{env}$
    \FOR{$k \in \{1,..,K\}$ synthetic rollouts per real episode}
        \STATE sample real episode $n' \sim \mathcal{U}\{1,\dots,n\}$
        \STATE sample step from episode $l_0 \sim \mathcal{U}\{0,T-L-1\}$
        \STATE set initial state $\tilde{s}_0^{(k)} = s_{l_0}^{(n')}$
        \FOR{$l \in \{0,\dots,L-1\}$ steps per synthetic rollout}
            \STATE sample action $\tilde{a}_{l}^{(k)} \sim \pi_\theta(a|\tilde{s}_{l}^{(k)})$
            \STATE building model returns $\tilde{s}_{l+1}^{(k)} = m_\psi(\tilde{s}_{l}^{(k)}, \tilde{a}_{l}^{(k)})$
            \STATE reward model returns $\tilde{r}_{l+1}^{(k)} = m_\beta(\tilde{s}_l^{(k)}, \tilde{a}_l^{(k)},\tilde{s}_{l+1}^{(k)}) $
        \ENDFOR
        
        \STATE $\tilde{D}_k \gets \{(\tilde{s}_0^{(k)},\tilde{a}_0^{(k)},\tilde{s}_1^{(k)},\tilde{r}_1^{(k)}),\dots,(\tilde{s}_{L-1}^{(k)},\tilde{a}_{L-1}^{(k)},\tilde{s}_L^{(k)},\tilde{r}_L^{(k)})\} $
        \STATE add synth. episode buffer to $D_{model} \gets D_{model} \cup \{\tilde{D}_k\}$
        
    \STATE apply RL to $\pi_\psi$ on experiences $D_{env} \cup D_{model}$
    \ENDFOR
\ENDFOR
\end{algorithmic}
\end{algorithm}

For example, Gao et al. \cite{gao2023comparative} focus on comparing model-based with model-free DRL, using an ensemble of 5 MLPs as surrogate model. They use two different DRL algorithms, DDQN, a deep Q-learning variant, and soft actor-critic (SAC). Their building MLP outputs a Gaussian distribution over states and rewards. At each time step, $K$ previously seen states are sampled from the measurement memory buffer. These are used as initial conditions for synthetic training, were the synthetic states are generated by the MLP ensemble. Their model-based versions converge 25\%-38\% faster than the model-free versions, but still need at least 46 weeks of environment interaction. Their training episodes also don't follow a chronological timeline. Furthermore, their model-based method \cite{janner2019trust} generates predictions for the entire state-space, e.g. the model also generates new weather and price forecasts. But these are invariant to control actions, and including them in the surrogate model prediction task increases the risk of model errors, which degrade RL training. Their training loop also allows the RL algorithm to time-travel, for example training one episode in April, then one in February, etc. 
Compared to them we use a counterfactual surrogate model instead of a general one, which makes the surrogate model more useful, and we train with episodes in a realistic chronological order. 

\cite{saeed2024dyna} compares the use of Dyna with a physics-informed neural network (Dyna-PINN), a standard neural network (Dyna-NN), and a resistor-capacitor model (Dyna-RC), using the DQN algorithm for reinforcement learning. Their PINN achieves around 1 degree Celsius mean absolute error (MAE) prediction performance (for 1-2 days forecast horizon), and their Dyna-PINN controller is particularly effective in low-diversity training data regimes. However, both their DRL agents and surrogate models still use 25-50 weeks of environment interaction by repeatedly sampling episodes from a 6-week window. Hence, they are also time-traveling. Their model rollouts are limited to only one to two days per week of real data. Furthermore, their state-space doesn't make use of forecasts and doesn't consider dynamic electricity prices. Compared to them, our counterfactual surrogate model can incorporate forecasts, makes use of dynamic prices and is trained on episodes in a chronological order.

\section{Methodology}
\label{sec:methodology}
In the following, we briefly cover notation, the basics of deep reinforcement learning and causal inference. Afterwards, we explain Dyna-style RL in Section~\ref{sec:our_dyna_style_mbrl} and our counterfactual surrogate model (CSM) in Section~\ref{sec:counterfactual_surrogate_model}. 

Throughout this work, we distinguish between real data, generated by interacting with the environment, and synthetic data, generated by the learned CSM $m$. The synthetic episodes using the CSM are also called model rollouts. The CSM contains a building model $m_\psi$ predicting zone temperatures and a reward model $m_\beta$ predicting rewards. We denote the state at time step $t$ within episode $n$ as $s_t^{(n)}$. Real variables are denoted with standard letters (e.g., $s, a, r$), while variables returned by the CSM are marked with a tilde (e.g., $\tilde{s}, \tilde{a}, \tilde{r}$). We index time in the real environment with $t$ and in model rollouts with $l$. We use $x \sim \mathbb{P}$ to denote sampling $x$ from a distribution $\mathbb{P}$. $\mathcal{U}\{1,\dots,n\}$ represents the uniform distribution over $\{1,\dots,n\}$. 




We formalize our problem as a Markov Decision Process (MDP) defined by the tuple $(\mathcal{S}, \mathcal{A}, P, R, \gamma)$, where $\mathcal{S}$ is the state space, $\mathcal{A}$ is the action space, $P(s_{t+1}|s_t, a_t)$ represents the transition dynamics, $R(s_t, a_t,, s_{t+1})$ is the reward function, and $\gamma \in [0, 1)$ is the discount factor.In Deep Reinforcement Learning (DRL), we aim to learn a policy $\pi_\theta(a|s)$, parameterized by $\theta$, that maximizes the expected cumulative episodic discounted reward for an episode of length $T$:
\begin{equation}
\label{eq:max_ep_rewards}
    J(\pi_\theta) = \mathbb{E}_{\pi_\theta} \left[ \sum_{t=0}^{T-1} \gamma^t r_{t+1} \right]. 
\end{equation}
where $r_{t+1} = R(s_t,a_t,s_t)$, $\gamma$ is a discount factor. We use a state-of-the-art RL algorithms, proximal policy optimization (PPO) and soft actor-critic (SAC), to maximize \cref{eq:max_ep_rewards} using the python library stable-baselines-3. 


Causal inference \cite{pearl2009causality} allows one to reason beyond statistical associations like $\mathbb{P}(s,a)$,  towards cause-effect relationships, e.g. $a$ influencing $s$: $a \rightarrow s$ by using causal graphs. This allows estimation of counterfactual questions like ``What would have happened to $s_{t+1}$ had $\tilde{a_t}$ been taken instead of $a_t$?'' For detailed explanations we refer to Pearl \cite{pearl2009causality}. 

\subsection{Dyna-style Model-based RL}
\label{sec:our_dyna_style_mbrl}
Dyna \cite{sutton1991dyna} is a model-based RL method that iteratively switches the policy $\pi_\theta$ between learning from the real environment $\mathbb{P}(s_{t+1}|s_t,a_t)$ and a surrogate model $m(s_{t+1}|s_t,a_t)$ with learnable parameters $\psi$. This allows $\pi_\theta$ to learn from large amounts of data from model rollouts using $m$, without discarding the real data from $\mathbb{P}$. 

\cref{alg:counter_dyna} illustrates our method. It runs for a total of $N$ episodes. Each of these is split up into the real environment interaction, lines 3-11 and synthetic environment interactions, which we also call model rollouts, in lines 12-21. 
The first part works like any model-free RL algorithm, learning after each episode from the  experiences $D_{env} \cup D_{model}$ that have been collected so far. 

At the end of each episode, the counterfactual surrogate model $m$ is fit on the collected real experiences $D_{env}$ such that it predicts the next states $s_{t+1}$. Further discussion on the model will follow in \cref{sec:counterfactual_surrogate_model}. The reward $r_{t+1} = R(s_t, a_t, s_{t+1})$ is calculated using the transitions $(s_t, a_t, s_{t+1})$, from a possibly unknown function. 


Per episode $n$, we generate $K$ model rollouts by first sampling an initial state from a real experience stored in $D_{env}$, and then letting $\pi_\theta$ and $m$ interact iteratively. The literature \cite{janner2019trust} suggests that shorter model rollouts $L << T$ are generally preferred. Long rollouts can suffer from accumulating model errors, which lead RL agents $\pi_\theta$ to learn wrong associations. After some rollouts, depending on the RL algorithm, the policy $\pi_\theta$ is updated, potentially several times during the model rollout phase. We repeat this combination of real episode and multiple simulated rollouts $N$ times. A visual overview of our training procedure can be seen in \cref{fig:dynambrl_visualized}. 

\subsection{Counterfactual Surrogate Model}
\label{sec:counterfactual_surrogate_model}
Dyna-style model-based RL relies on a surrogate model $m$ that can generate accurate and robust predictions with little data. It's prediction task should be as easy as possible. Our counterfactual surrogate model contains a building model $m_\psi$ (with learnable parameters $\psi$), which predicts zone temperatures, and a reward model $m_\beta$ (with parameters $\beta$), which predicts rewards. 

\textbf{Building model. }In contrast to previous work, which uses the surrogate model to predict the \emph{entire} next state $s_{t+1}$ \cite{janner2019trust,gao2023comparative}, or doesn't use forecasts \cite{saeed2024dyna}, we split the state-space $s_t = (z_t,d_t)$ into the zone (building) temperature $z_t$ and disturbances/forecasts $d_t$, which contain weather forecasts, price forecasts and time. This allows us to fit the surrogate model (containing the building model and reward model) only on the variables that will change ($z_{t+1}$) for alternative action sequences. Weather forecasts, price forecasts and time can reasonably be assumed to be not influenced by heat pump activations, so they do not need to be estimated, likely erroneously, in the surrogate model rollouts. This simplifies the CSM's task. 

\begin{figure}
    \centering
    \includegraphics[width=0.48\textwidth]{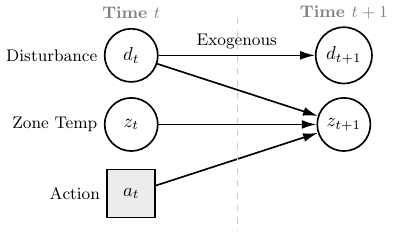}
    \caption{Causal diagram describing the causal relationships between actions and state-space variables.}
    \label{fig:counterfactual_env_graph}
\end{figure}


A causal graph describing the relationship of the variables $z_t$, $d_t$, $a_t$, $z_{t+1}$ and $d_{t+1}$ is shown in \cref{fig:counterfactual_env_graph}. Using causal inference \cite{pearl2009causality} to reason about the effect of intervening on actions and thus indirectly on zone temperature (denoted by the $do$-operator), the next state disturbances in a model rollout $\tilde{d}_{l+1}$ aren't influenced by $(\tilde{z}_l,\tilde{a}_l)$:
\begin{align}
\label{eq:disturbance_invariance_interventional}
    \mathbb{P}(\tilde{d}_{l+1}|\tilde{d}_{l},do(\tilde{z}_t), do(\tilde{a}^{(k)}_t))) 
    = \mathbb{P}(\tilde{d}_{l+1}|\tilde{d}_{l})
\end{align}
This means that for a synthetic model rollout, the observed historical disturbances $d_0,...,d_T$ can be used and don't need to be estimated. The  state $\tilde{s}_{l+1}^{(k)}$ in a model rollout can thus be written as 
\begin{align}
    \tilde{s}_{l+1}^{(k)} = (\tilde{z}_{l+1}^{(k)}, \tilde{d}_{l+1}^{(k)}) = (\tilde{z}_{l+1}^{(k)}, d_{l_0+l+1}^{(n')}), 
\end{align}
where the model rollout disturbances $\tilde{d}_{l+1}^{(k)}$ can be replaced with the real disturbances $d_{l_0+1}^{(n')}$ from episode $n'$ that synthetic episode $k$ was sampled from. 
Thus it is enough to fit a building model to only forecast the next zone temperature: 
\begin{align*}
    \tilde{z}_{l+1}^{(k)} = m_\psi(\tilde{z}_{l}^{(k)}, d_{l_0 + l}^{(n')}, \tilde{a}_l^{(k)})
\end{align*}
The first step in a model rollout $k$ can be seen as a conditional interventional prediction \cite{pearl2009causality} of what would have happened in state $s_{l_0}^{(n')}$, if instead of action $a_{l_0}^{(n')}$, action $a_{0}^{(k)}$ would have been taken:
\begin{align*}
    \tilde{z}_{1}^{(k)} = m_\psi\!\left(z_{l_0}^{(n')}, d_{l_0}^{(n')}, \tilde{a}_0^{(k)}\right) \approx \mathbb{E} \left[ Z_{l_0+1} \mid S_{l_0} = s_{l_0}^{(n')}, \operatorname{do}(A_{l_0} = \tilde{a}_{0}^{(k)}) \right]
\end{align*}
This holds for the first state $l = 1$, and further states $l > 1$ are then a forward propagation of a sequence of alternative actions impacting this counterfactual world:
\begin{align}
\label{eq:counterfactual_rollout}
    \mathbb{E} [ \tilde{Z}_{l_0+l+1} \mid \tilde{Z}_{0}^{(k)} = z_{l_0}^{(n')}, d_{l_0:l_0+l}, \text{do}(A_{l_0:l_0+l} = \tilde{a}_{0:l}^{(k)}) ]
\end{align}

\cref{eq:counterfactual_rollout} admits a clear counterfactual interpretation: It is the expectation of the zone temperature at time $l_0 + l + 1$ if an alternative action sequence $\tilde{a}_{0:l}^{(k)}$ had been taken. This causal reasoning reduces the dimension to be predicted from $dim(s)$ to $dim(z)=1$, so the forecasting problem becomes easier and thus more accurate. 
In addition, since the model rollouts are using historical data, the disturbances $\tilde{d}_l^{(k)}$ in state $s_l^{(k)} = (\tilde{z}_l^{(k)}, \tilde{d}_l^{(k)})$ are also perfectly accurate. Put together, this makes our model rollouts significantly more precise, which enhances the simulated experience for the RL agent. Additionally, this reliability allows us to generate more rollouts, as they aren't so likely to degrade the RL agent's performance, which speeds up its convergence (measured in real interactions). Instead of predicting zone temperatures directly, our MLP predicts the difference between the current and next step zone temperatures. 


\textbf{Reward model. }Forecasting the next state is not enough to obtain a complete surrogate model. It also has to return the reward. In BOPTEST, the control objective is defined via two KPIs: thermal discomfort and HVAC operational cost. Thermal discomfort over a time period $[t_0,t_f]$ is defined as the zone-average time integral of the comfort-band slack $\delta(u)$,
\begin{align}
    R_{tdis}(t_0,t_f) = \frac{1}{N}\int_{t_0}^{t_f} \left|\delta(t)\right|\,dt,
\end{align}
where $\delta(u)$ denotes the deviation of the zone temperature from the predefined comfort range (e.g. $[21\text{°C},24\text{°C}]$) at time $t$. The unit of $R_{tdis}$ is Kelvin-hours. For example, one Kelvin hour can be a comfort violation of 1°C for 1 hour, or 0.5°C for 2 hours.

Operational cost is defined as the floor-area-normalized integral of an electricity price $\tau(t)$ (in €/kWh) times power across all HVAC energy vectors/equipment $i\in\xi$,
\begin{align}
    R_{cost}^{\tau}(t_0,t_f) = \frac{1}{A}\sum_{i\in\xi}\int_{t_0}^{t_f} \tau(t)\,P_i(t)\,dt,
\end{align}
where, $P_i(t)$ is the corresponding power consumption, and $A$ is the total floor area. The unit of $R_{tdis}$ is in € per square meter, €/$m^2$. 

The reward is the negative weighted sum of discomfort and cost,
\begin{align}
    r_{t+1} = -(w_D R_{tdis}(t,t+1) + w_C R_{cost}^\tau (t,t+1))
\end{align}
with nonnegative weights $w_D$ and $w_C$ determined to balance a tradeoff between cost and comfort. We use $w_D = 1, w_C = 100$ for our experiments, as the cost term has significantly lower scale. \cref{sec:ablation_study} gives more details about the impact of varying $w_C$. $w_D = 1, w_C = 100$ showed to be a good balance. 

While BOPTEST can return equipment power usage $P_i(t)$, we are not using this data to make our setup more general. 
    The CSM thus also doesn't model it, and hence we can't directly compute the energy cost part $R_{cost}^\tau(t,t+1)$ of the reward. Therefore, we create a piecewise linear reward model $m_\beta(a_t, \tau_t, a_t\tau)$ that is tasked with predicting the reward based on state-space data that is available, which is the heat pump activation $a_t$, prices $\tau(t)$ and their product $a_t\tau$. First, when the heat pump is off, $a_t = 0$, all equipment is switched off, so $m_\beta(0,\tau,0) != 0$. For non-zero activations, we use a linear model with parameters $\beta = (\beta_0,\beta_1,\beta_2,\beta_3)$
which gives
\begin{align*}
    m_\beta\!\left(a_t,\tau_t,a_t\tau_t\right)=
\begin{cases}
0, & a_t = 0\\
\beta_0 + \beta_1 \,a_t + \beta_2\,\tau_t + \beta_3\,(a_t\tau_t)&  a_t\neq 0
\end{cases}
\end{align*}
The parameters $\beta$ are fit using least squares regression based on the collected real data $D_{env}$.

\section{Evaluation}
\label{sec:evaluation}

In this section we illustrate our evaluation. We first explain our experimental setup in \cref{sec:experimental_setup}. Then we show the performance of Counter-Dyna compared to benchmark methods and the accuracy of the CSM in \cref{sec:results}. \cref{sec:discussion} discusses our results and \cref{sec:limitations} mentions the limitations of our work. 



\subsection{Experimental Setup}
\label{sec:experimental_setup}
We use the bestest hydronic heat pump testcase from BOPTEST \cite{blum2021building}, which has become a standard for evaluating RL algorithms in BOPTEST \cite{wang2023comparison,gao2023comparative,saeed2024dyna,WANG2025124328}. We added more recent electricity market prices for 2024 from Belgium (compared to previous prices of 2019, Belgium), which are even more volatile than the previous highly dynamic pricing scheme. This makes the optimization problem more challenging, but also more realistic. The new prices reflect changes that influenced electricity markets in recent years due to an energy crisis and the increasing share of renewables. We publish our code under \codelink, which makes our results verifiable and reproducible. 

\textbf{Experiment Setup. }We focus our evaluation on four experiments. Each experiment is repeated 30 times with different random seeds to ensure statistically robust results. Within each experiment, RL agents using the PPO algorithm \cite{schulman2017proximal} with two discrete actions (heat pump on or off) were trained. The goal is to determine the data efficiency of Counter-Dyna compared to model-free and baseline benchmarks, and their control performance. 

For this, in two experiments, model-based agents using Counter-Dyna are trained. One experiment trains for only 5 weeks, and another trains the policy for 10 weeks. We call them Counter-Dyna-5 and Counter-Dyna-10, respectively. Since 1 episode is 1 week long, we sometimes use the the two terms interchangeably. Two further experiments employing model-free (MF) algorithms serve as comparison. In the model-free runs, one trains for 10 weeks and another for 50 weeks. We call them MF-10 and MF-50, respectively. The 10-week Counter-Dyna and MF runs are equal in terms of training data and it can be directly compared whether using Counter-Dyna or MF is preferable. The 5-week Counter-Dyna runs show what is possible for Counter-Dyna using even less time. The 50-week MF runs show the upper limit for a model-free approach, as the testcase only has one year of data, where the remaining 2 weeks serve for testing. It is important to note that the runs not only differ w.r.t. their training amount, but also when their training was started. This is due to our chronological episodes ending just before January 17th, so that they can be tested on the peak heating period from Jan 17th - Jan 31st. Different training lengths thus result in different start times. The hyperparameters for the RL algorithms and the CSMs are listed in \cref{sec:hyperparameter_tables}. We tuned them manually, varying one at a time based on educated guesses. 

We performed two additional studies. One used the SAC algorithm \cite{haarnoja2018soft} with continuous actions in the interval $[0,1]$ instead of PPO with discrete actions. Since the results are similar to the PPO runs, we show these results in less detail in \cref{sec:comparing_dyna_mf_sac}. This study examines the robustness of Counter-Dyna to RL algorithm choice and action space definition. We also did an ablation study, investigating the influence of the amount of synthetic data, rollout length and cost influence on reward $w_C$ in \cref{sec:ablation_study}. 


\textbf{Metrics. }BOPTEST computes several key performance indicators (KPIs), of which we use the discomfort and energy cost (see \cref{sec:counterfactual_surrogate_model}) to assess the performance of our controllers. We also compare these metrics w.r.t. the number of episodes used to train an RL agent, as convergence speed is important for real-world applications. 

\textbf{Benchmarks \& Hyperparameters. }We use the BOPTEST baseline PI controller and a rule-based controller as further benchmarks beyond the model-free versions. The rule-based controller has a 60min control step and tracks a temperature of 21.5°C, turning the heat pump on (off) when below (above) the threshold. This bang-bang control always operates at high COP when the heat pump is running, making this controller particularly cost-effective. However, this will not hold in the real world, as BOPTEST doesn't enforce minimum run-times or start-up losses \cite{blum2021building}, where modulating controls are expected to be more efficient and suitable for reducing equipment wear. The state space given to the control agent and building model is described in \cref{tab:state_space}. 

\textbf{Test periods. } BOPTEST defines two standard testing periods, the peak heating period and typical heating period. The peak heating period are the two coldest weeks of the year, going from January 17th to January 31st. The typical heating period is from April 19th to May 3rd. Since we train our agents in an online fashion with chronological episodes, we are setting them up so that their training ends just before the peak heating period. The 5-week Counter-Dyna and 10-week Counter-Dyna and MF only observed data in the winter before January 17th. This means that for these runs, the typical heating period measures their generalization ability to a different spring weather regime. 

\begin{table}[htpb]
\caption{State space for the discrete PPO algorithm. The total dimension is $4 + 1 + 7 + 7= 19$. }
\centering
\begin{tabular}{p{1.3cm} p{1.3cm} p{4.3cm}} 
\textbf{Variable} & \textbf{Value/Unit} & \textbf{Description} \\ \hline
time & $[-1,1]^{4}$ & Cyclical encoding based on hour and weekday. \\ \hline
reaTZon\_y & K & Zone temperature in Kelvin. \\ \hline
TDryBul & K & Ambient temp. in K, forecast horizon 6h. \\ \hline
price & €/kWh & Elec. price, forecast horizon 6h. \\ 
\end{tabular}
\label{tab:state_space}
\end{table}


\subsection{Results}
\label{sec:results}



\begin{figure*}[!ht]
    \centering
    \includegraphics[width=0.95\textwidth]{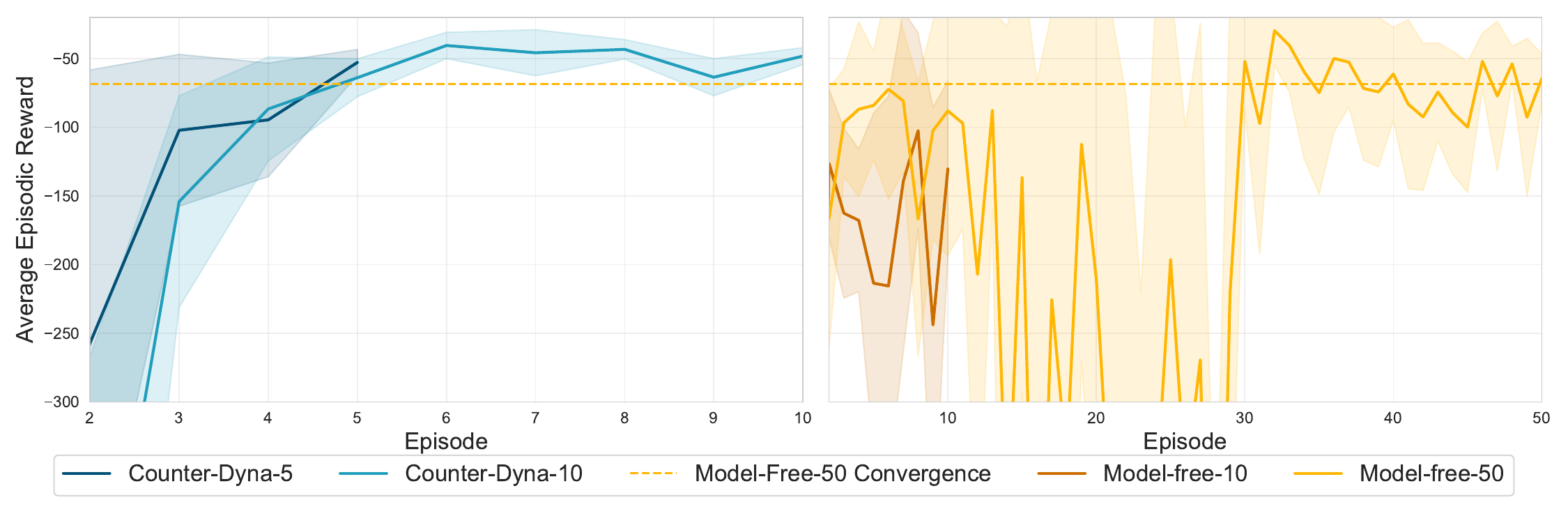}
    \caption{Average episodic rewards. 1 week = 1 episode. \textbf{Left: } Counter-Dyna runs for 5 and 10 weeks. \textbf{Right: Model-free runs for 10 and 50 weeks. Shaded regions indicate standard deviations across 30 seeds. The 50-week MF asymptotic performance is shown as a dashed line, it is the average of its episodes 30-50. }}
    \label{fig:convergence_comparison}
\end{figure*}

We first analyze the convergence of the 4 experiments, which are visualized in \cref{fig:convergence_comparison}. Both Counter-Dyna-5 and Counter-Dyna-10 show quick performance, where Counter-Dyna-10 seems to converge at episode 6. Counter-Dyna-5 doesn't show convergence yet, however both Counter-Dyna-5 and Counter-Dyna-10 surpass the asymptotic MF-50 after only 5 weeks. Both experiments are also quite stable and robust to random seeds, indicated by the rapid reduction of the shaded areas which indicate the standard deviation. MF-10 and MF-50 show lower and much more volatile rewards. MF-10 shows no sign of convergence. MF-50 converges after 30 weeks, which is 6 times longer than what it took both Counter-Dynas to achieve superior performance. 

\begin{figure*}[!ht]
    \centering
    \includegraphics[width=0.95\textwidth]{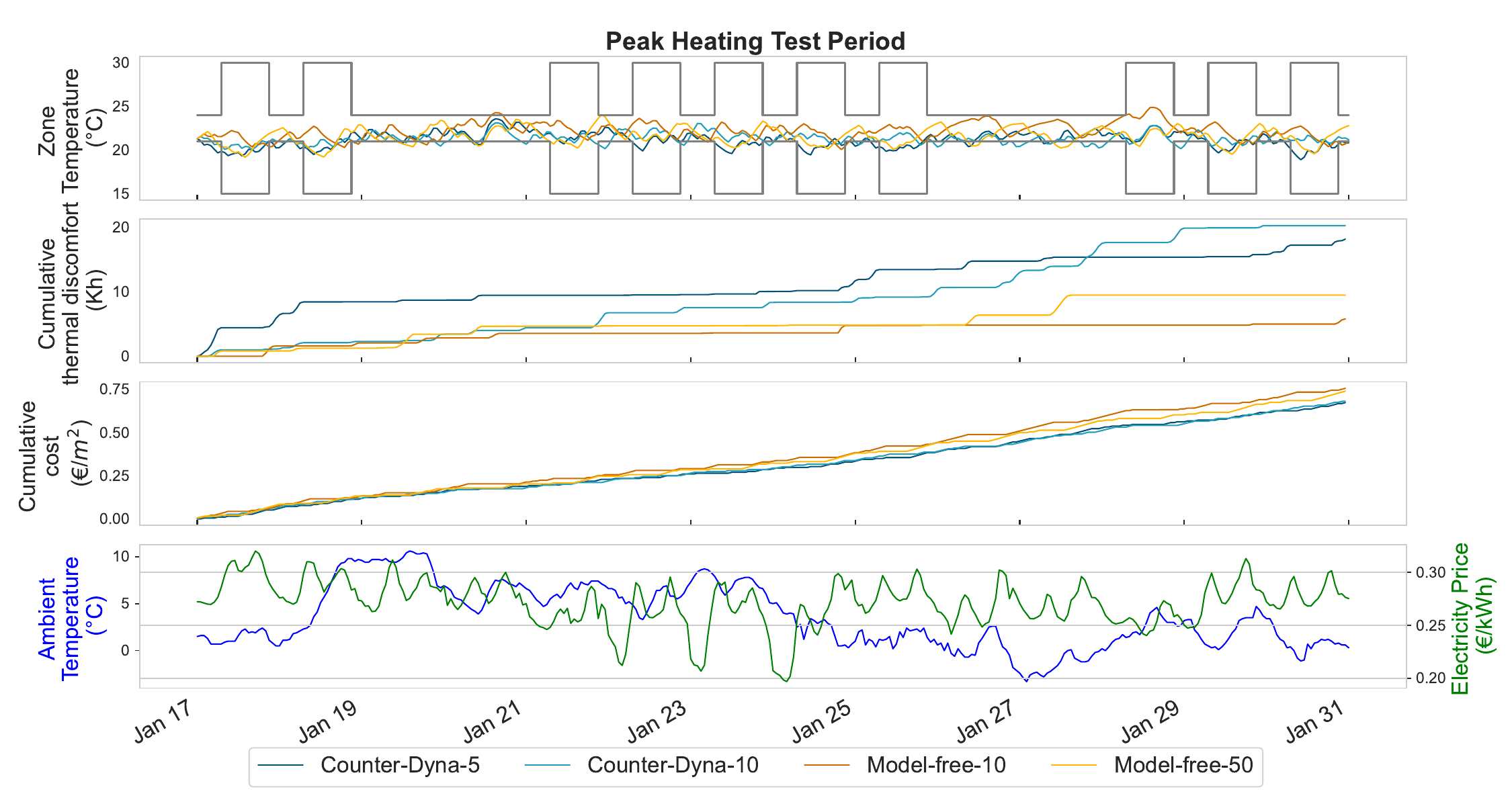}    \caption{Control performance of the best runs per category during the peak heating period.  }\label{fig:test_peak_heat_comparison_temperature_srewards_forecasts_all_runs}
\end{figure*}

\begin{figure*}[!ht]
    \centering
    \includegraphics[width=0.95\textwidth]{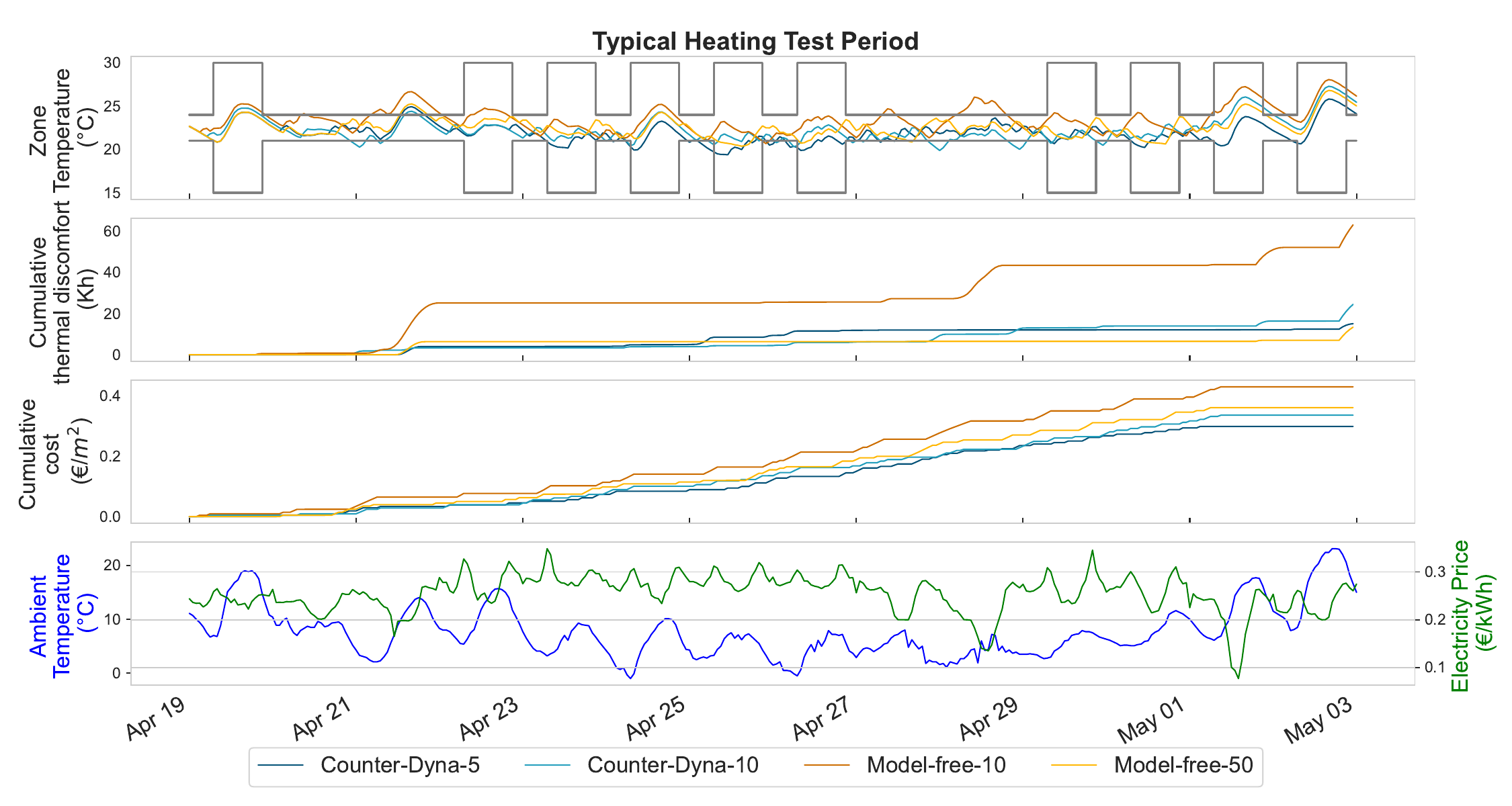}
    \caption{Control performance of the best runs per category during the typical heating period. }\label{fig:test_typ_heat_comparison_temperature_rewards_forecasts_all_runs}
\end{figure*}

Next, we examine the control performance. For this, we chose the 4 best runs for each category, from the 30 available runs per category. We chose runs that had the lowest cumulative cost, given that their cumulative thermal discomfort in the peak heating period was below 30 Kelvin-hours. This is lower than an average comfort violation of 0.1°C during the two week period. The peak heating period is shown in \cref{fig:test_peak_heat_comparison_temperature_srewards_forecasts_all_runs}. In the plot of the zone temperatures, all runs show good performance, with no large or long comfort violations visible. The cumulative thermal discomfort reveals a first difference. MF-10 and MF-50 achieve lower discomfort below 10, while Counter-Dyna-5 and Counter-Dyna-10 are slightly higher, near 20. However they are more costly, as the next plot shows. A different picture emerges for the peak heating period in \cref{fig:test_typ_heat_comparison_temperature_rewards_forecasts_all_runs}. The control performance for both Counter-Dynas and MF-10 is satisfactory, but MF-10 shows repeated comfort violations. This leads to MF-10 having significantly higher cumulative discomfort, while Counter-Dyna-5,-10 and MF-50 are close. Again, the Counter-Dyna runs are more cost-effective. 

\begin{figure}[ht]
    \centering\includegraphics[width=0.48\textwidth]{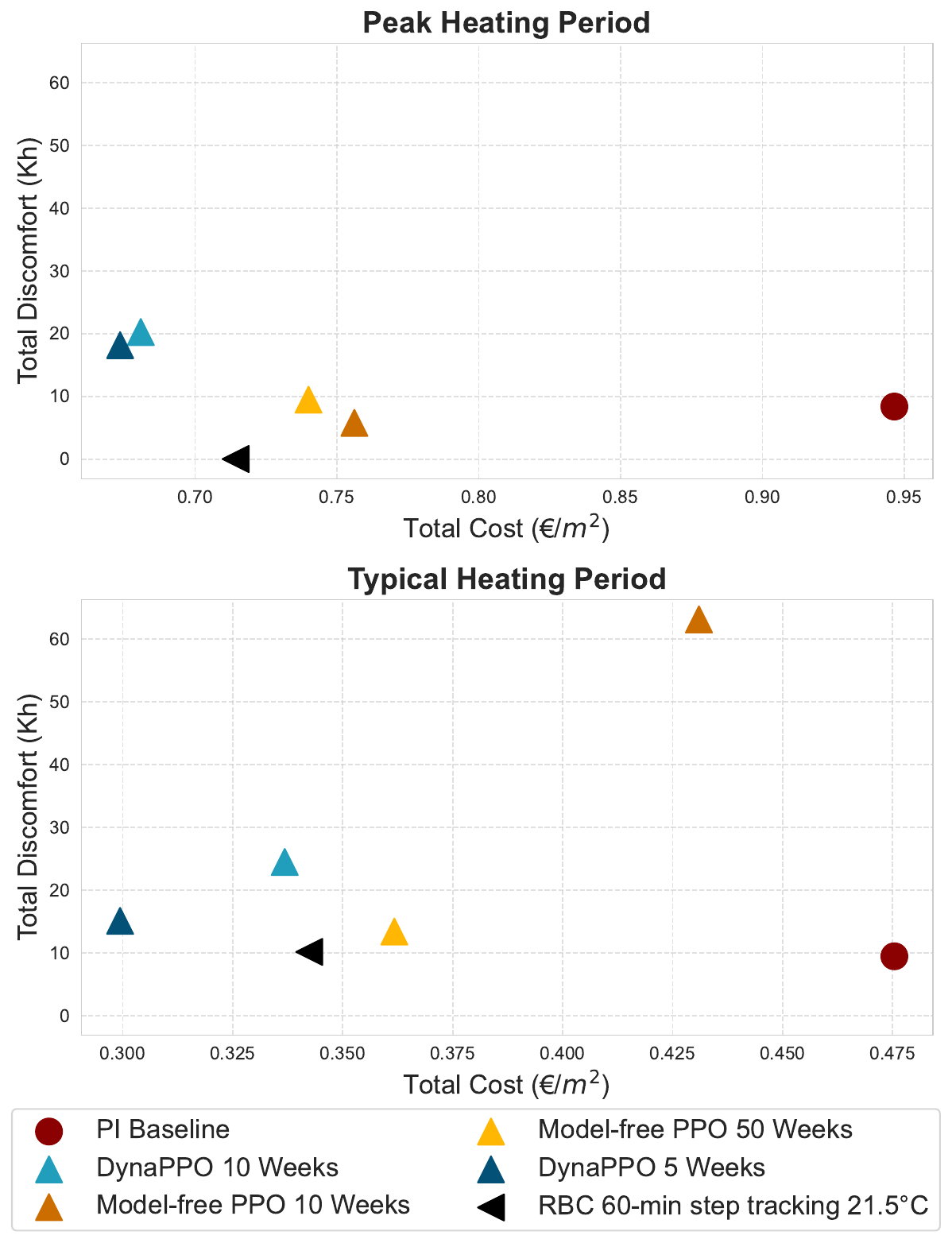}
    \caption{Scatterplot of cost and discomfort for the best runs for each setting. }
    \label{fig:cost_vs_tdis}
\end{figure}

The same best-performing runs are also compared in a scatterplot of discomfort versus cost in \cref{fig:cost_vs_tdis} to show a cost-discomfort tradeoff. With only minor discomfort violations, both Counter-Dynas achieve lower cost than the MF variants. The Counter-Dynas are also more cost-effective than the rule-based controller, which the MF variants are not. This holds for both testing periods. All RL algorithms are much cheaper than the PI baseline controller. 

\begin{figure*}[t]
    \centering
    
    \begin{minipage}{\textwidth}
        \centering
        \includegraphics[width=0.9\textwidth]{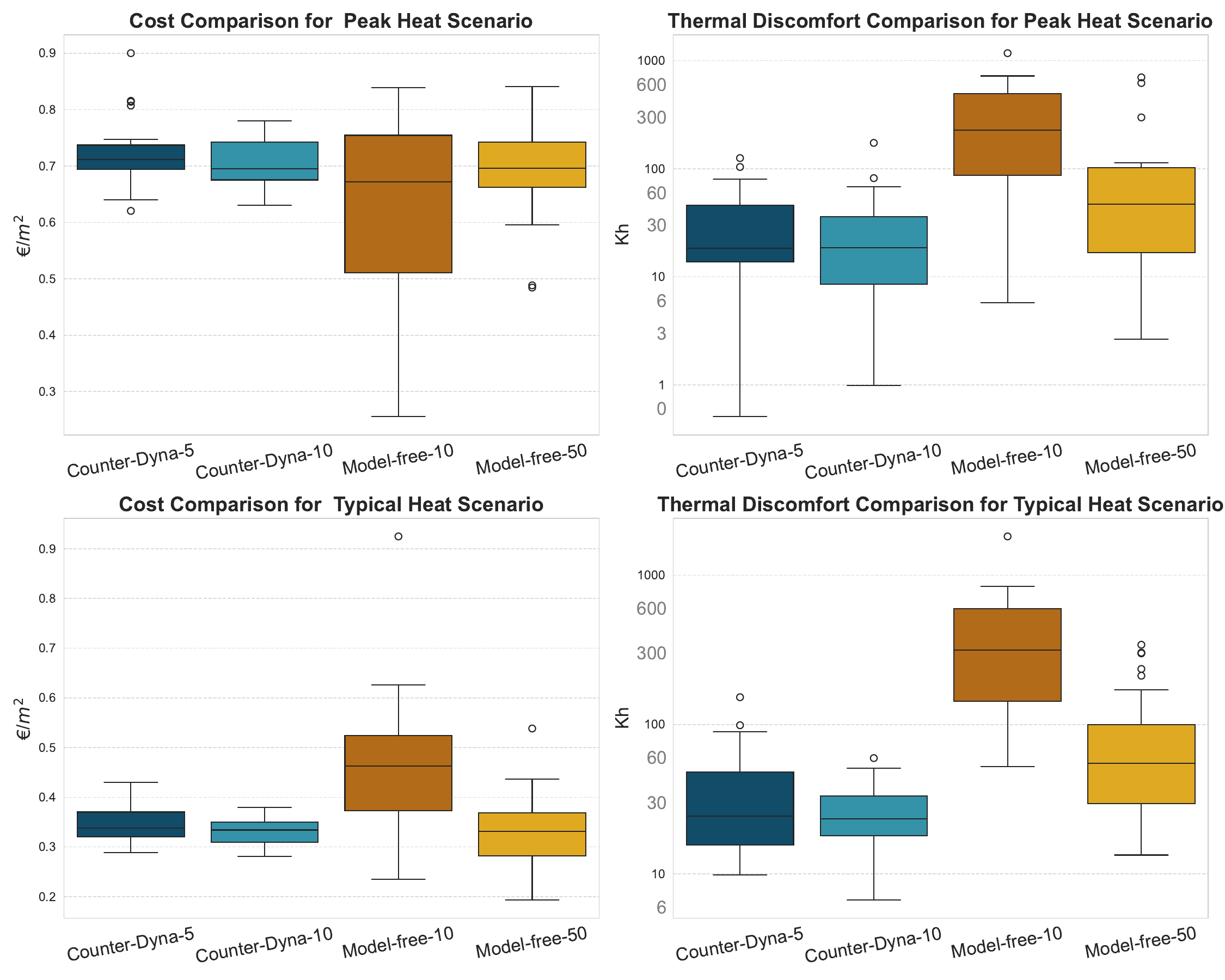}
        \caption{PPO box plots of cost and discomfort for both testing periods, with distributions over the 30 different seeds.}
        \label{fig:main_study_kpi_boxplots}
    \end{minipage}

    \vspace{2em} 

    \begin{minipage}{0.48\textwidth}
        \centering
        \includegraphics[width=\textwidth]{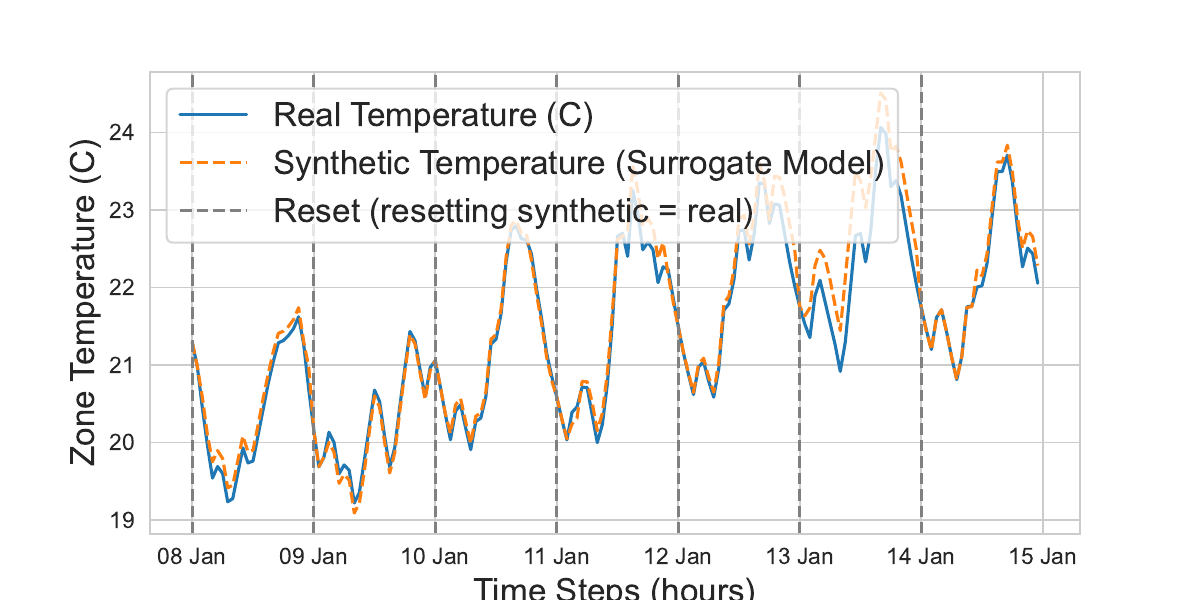}
        \includegraphics[width=\textwidth]{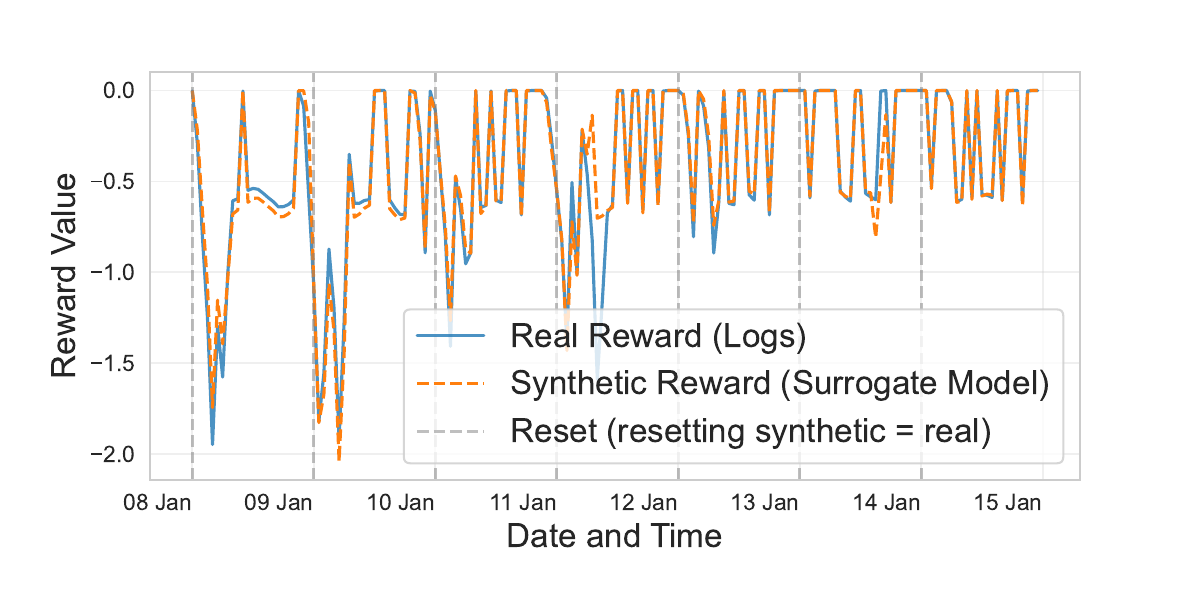}
        \caption{\textbf{Top: }Comparing zone temperatures for in-sample actions. \textbf{Bottom: } Corresponding rewards.}
        \label{fig:surrogate_model_in_sample_actions}
    \end{minipage}
    \hfill
    \begin{minipage}{0.48\textwidth}
        \centering
        \includegraphics[width=\textwidth]{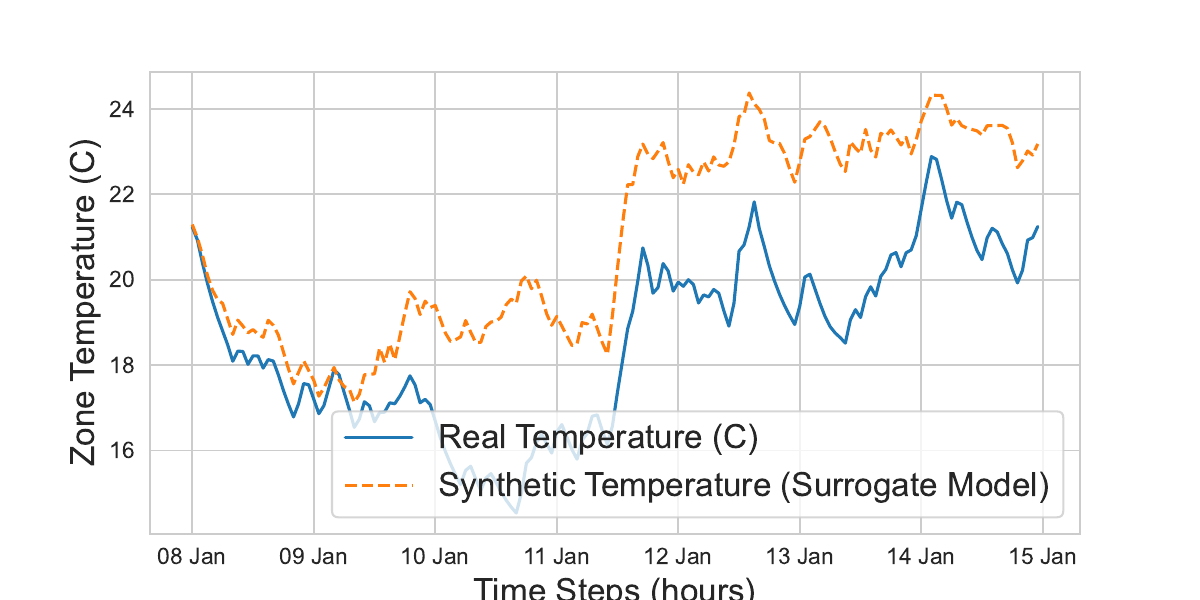}
        \includegraphics[width=\textwidth]{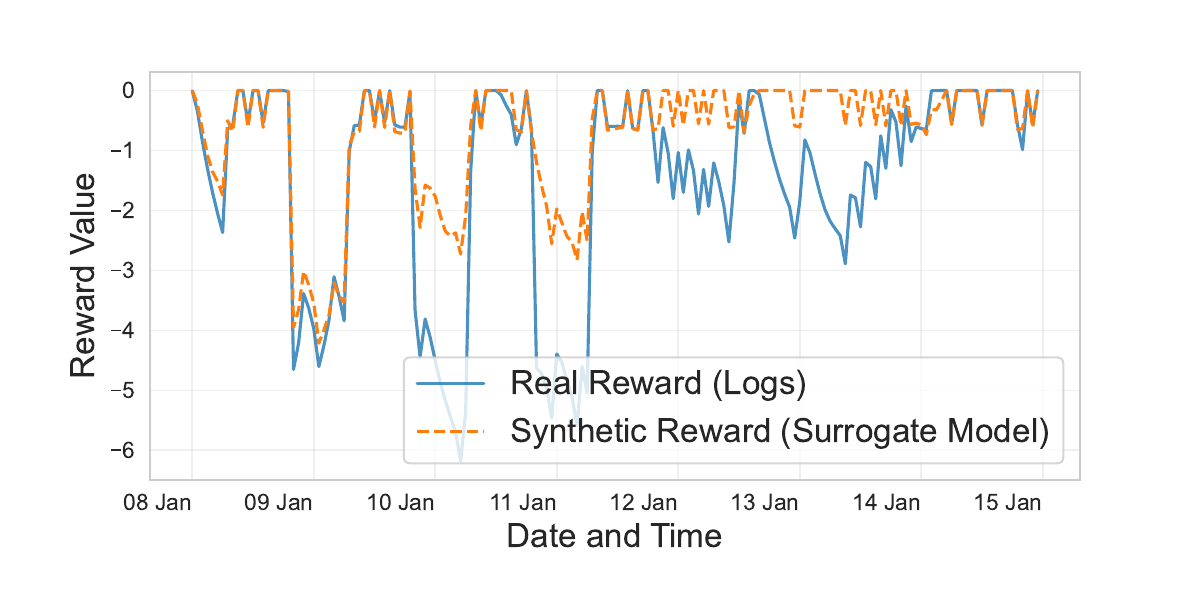}
        \caption{\textbf{Top: }Comparing zone temperatures for out-sample actions. \textbf{Bottom: } Corresponding rewards.}
        \label{fig:surrogate_model_out_of_sample_actions}
    \end{minipage}
\end{figure*}

The box plots in \cref{fig:main_study_kpi_boxplots} show the distribution of cost and discomfort for both testing periods. Regarding cost during peak heating, Counter-Dyna-5 and Counter-Dyna-10 show lower average and lower variation compared to MF-50. MF-10 has a lower average cost, but this is due to its failure to respect comfort bounds, note the log scale on the y-axis here. Counter-Dyna also have lower average discomfort and lower variation than MF-50 regarding discomfort. The typical heating period is similar, except that the comfort violations of MF-10 are paired with higher average costs, indicating poor generalization performance. Counter-Dyna is again better and more stable than MF-50. Counter-Dyna-10 and MF-10 are equivalent in terms of training time, so their cost-efficiency can be compared for a hypothetical deployment scenario. For runs that had overall discomfort below 100Kh, Counter-Dyna-10 achieves cost savings of 5.3\% in the peak heating period and
17.0\% in the typical heating period. Counter-Dyna-5 is also more cost-efficient than MF-10, with 3.5\% and 13.2\% savings respectively, with using only half the training time. MF-50 had similar cost as Counter-Dyna-5 and Counter-Dyna-10, but with higher discomfort.

Next we evaluate the CSM, which contains the building model forecasting zone temperatures, and a reward model, which estimates rewards. We compare the building model with the BOPTEST environment for in-sample and out-of-sample actions. In-sample actions are those that the agent used when interacting with the BOPTEST environment, i.e. these are in the training data. Out-of-sample actions ask the building model to answer counterfactual queries, i.e extrapolating:``What would have happened to the zone temperature had $a=0$ been taken instead of $a=1$?''. We evaluate the accuracy of the building model on these counterfactual queries by initializing both the model and BOPTEST at the same operating point and then simulating them forward under a new action sequence. We measure the building models accuracy using root mean squared error (RMSE) and mean absolute error (MAE) in \cref{tab:metrics_bm}. 

\begin{table}[h]
\caption{Average building model temperature accuracies (mean $\pm$ SD) for 1-day rollouts for the final building models of the 5-week Dyna runs. }
\label{tab:metrics_bm}
\centering
\begin{tabular}{lcc}
\textbf{Metric} & \textbf{In-sample Actions} & \textbf{Out-of-sample Actions} \\ \hline
RMSE            & 0.08 $\pm$ 0.04 °C          & 0.90 $\pm$ 0.84 °C          \\ \hline
MAE             & 0.19 $\pm$ 0.05°C          & 1.66 $\pm$ 2.77 °C          \\ 
\end{tabular}
\end{table}

Over the lifetime of Counter-Dyna, multiple CSM versions exist, as it is retrained every episode $n$. Here we evaluate the \emph{final} CSM that resulted from the 5-week Counter-DynaPPO runs. The in-sample results are shown in \cref{fig:surrogate_model_in_sample_actions} for a 7-day period. We can see that the building model is very accurate, even for 24 ahead forecasts, where its own prediction is fed back into the model for the next forecast. Every 24h, the surrogate model is reset to the real buildings' state, as our model rollouts are also 24 steps. The reward model is also very accurate, having an $R^2$-score of 0.98, which means 98\% of the reward variation is captured by the reward model. It deviates only from the true reward if the building model deviates. 




For out-of-sample action sequences, the building model is less accurate as can be seen in \cref{tab:metrics_bm}. It however still captures the general building dynamics well, see \cref{fig:surrogate_model_out_of_sample_actions}. This is good enough for the RL agents to learn from CSM rollouts. Again the reward model is very accurate with an $R^2$-score of 0.98 and only deviates from the true reward when the building model is inaccurate. 



\subsection{Discussion}
\label{sec:discussion}

We evaluated two Counter-Dyna algorithms, training for 5 and 10 weeks, respectively, with two model-free algorithms, training for 10 and 50 weeks, respectively. We compared them in a realistic deployment scenario with chronological episodes using the BOTPEST bestest hydronic heatpump testcase. 

Our results show that Counter-Dyna-5 achieves stronger performance than MF-50 in a tenth of the time. Counter-Dyna also shows remarkable stability in terms of control performance, both w.r.t. different initializations and randomizations of the training algorithm. The same is true for its generalization performance on an unseen weather regime. Counter-Dyna-5 and 10 outperform MF-50, which has seen various weather regimes. Counter-Dyna decisively outperforms MF-10, which fails for the typical heating period. 

The slight comfort violations of Counter-Dyna show that it follows its control objective very well, which by design trades off some discomfort against cost. 

The reason for Counter-Dynas effectiveness lies in the accuracy of the counterfactual surrogate model, which contains the building and reward model. We showed that even for out-of-sample action sequences, the building model captures the building dynamics well. The reward model near-perfectly predicts rewards. A good reward model is very important, as even when the building model simulates biased states, the reward still gives the correct signal based on that state. All this makes the CSM a high-quality surrogate training environment, even during very early training and gives Counter-Dyna ample opportunity to learn ``between'' real episodes. This explains its strong performance, robustness and generalization ability. 




\subsection{Limitations}
\label{sec:limitations}
Despite a significant step towards practically relevant convergence speeds, our study has several limitations. While using BOPTEST is a standard way of evaluating control algorithms, the testcase is still a simplification compared to deployment in a real environment. 

Our setup uses perfect forecasts, which are not available in practice. Incorporating inaccurate forecasts into Counter-Dyna will need thought on whether how to combine historical forecasts with realized observations in the CSM. We also only use one architecture, an MLP, to predict future zone temperatures. Analyzing the interplay of various building models with Counter-Dyna could be interesting future work. 

And even though we train RL agents in an online fashion, we are still resetting the BOPTEST testcase at the episode boundaries, jumping to a completely new operating point. In practice, this could be addressed by letting a baseline controller run for a few hours, 'resetting' the environment for the next episode of RL. In addition, to safeguard against large constraint violations, safety constraints could be added. 

Additionally, Transfer learning could speed up the surrogate model’s initial training, followed by continuous learning techniques to update the model progressively over time, as in \cite{RAISCH2026116868}.

\section{Conclusion}
\label{sec:conclusion}

In this paper we introduced Counter-Dyna, a novel model-based reinforcement learning method for HVAC control. By redesigning and simplifying the model prediction task, we create a counterfactual surrogate model that serves as an additional RL training environment. We validate our method in an extensive simulation study using BOPTEST, and do so using realistic chronological episodes. Our experiments show that Counter-Dyna speeds up RL training w.r.t. real environment interactions by a factor of 10, and achieves superior performance compared to model-free benchmarks. In a hypothetical deployment scenario, Counter-Dyna can save costs of 5.3\% to 17.0\% while achieving lower thermal discomfort. Counter-Dyna's training is also robust and it generalizes well to an unseen weather regime. Our work is, to our knowledge, the first that, by model-based Dyna-style augmentation, speeds up a model-free RL algorithm like PPO or SAC to be deployable in just a few weeks of training. 




\clearpage
\begin{acks}
We thank our colleague Xiwen Huang from Imperial College London for the fruitful discussions during this project. 
\end{acks}

\balance
\bibliographystyle{ACM-Reference-Format}
\bibliography{bibliography}

\appendix
\crefalias{section}{appendix}

\section{Comparison of Counter-Dyna and Model-free Soft-Actor-Critic with Continuous Actions}
\label{sec:comparing_dyna_mf_sac}

To show our method for more than one RL algorithm and a different action-space, we train SAC algorithms in the same fashion as described in \cref{sec:methodology,sec:evaluation}. SAC works with a continuous action space, so the heat pump signal can be any smooth activation between 0 and 1. We use 20 seeds in this study. The hyperparameters used can be found in \cref{tab:sac_hyperparameters}. We didn't perform as extensive hyperparameter tuning in this setup, so we expect results to be overall less strong than the PPO runs. 

The box plots show that overall, SAC is significantly less cost-efficient than PPO. This is expected, because in this BOPTEST testcase, bang-bang control is more efficient than modulating control, due to modeling simplifications. In the peak heating period, Counter-Dyna has slightly higher cost, but very low discomfort. In particular the MF-50 runs have unacceptably high discomfort on average. In the typical heating scenario, MF-10 fails with high cost and discomfort, while Counter-Dyna-10 performs the best in both categories. Counter-Dyna-5 as slightly higher cost and discomfort than MF-50. 

\begin{figure*}[b]
    \centering
    \includegraphics[width=0.9\textwidth]{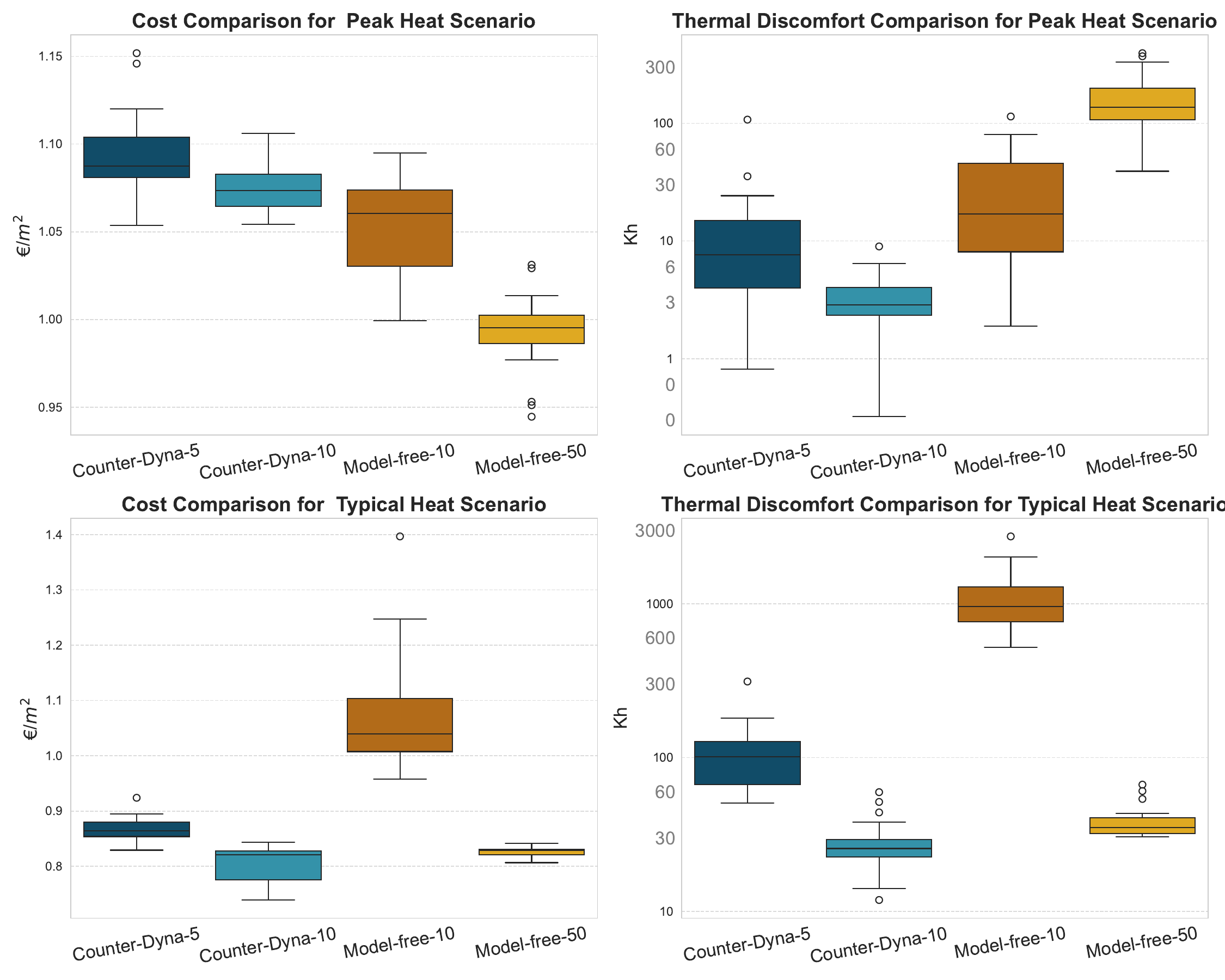}
    \caption{SAC box plots of cost and discomfort for both testing periods, with distributions over the 30 different seeds. }
    \label{fig:main_study_kpi_boxplots_sac}
\end{figure*}

\clearpage
\newpage
\section{Ablation Study of Dyna Hyperparameters}
\label{sec:ablation_study}

Dyna-style RL introduces two key hyperparameters: (i) how much synthetic experience is generated and (ii) the rollout length of each synthetic episode. We quantify the synthetic-data budget as a synth:real ratio, i.e., the amount of model-generated data consumed per unit of real data. For a training cycle with $K$ synthetic episodes of length $L$ and real episode length $T$, this ratio is
\[
\frac{KL}{T},
\]
where $K$ is the number of synthetic episodes, $L$ is the rollout length per synthetic episode, and $T$ is the length of a real episode.

To study sensitivity to these choices, we run an ablation where we vary one hyperparameter at a time while holding the other at its baseline value (synth:real ratio $=20$ and rollout length $L=24$). Each configuration is trained for 5 episodes (1 episode = 1 week) and repeated three times with different random seeds; the evaluated settings are listed in \cref{tab:ablation_study_hparams}.

\begin{figure}
    \centering
    \includegraphics[width=0.48\textwidth]{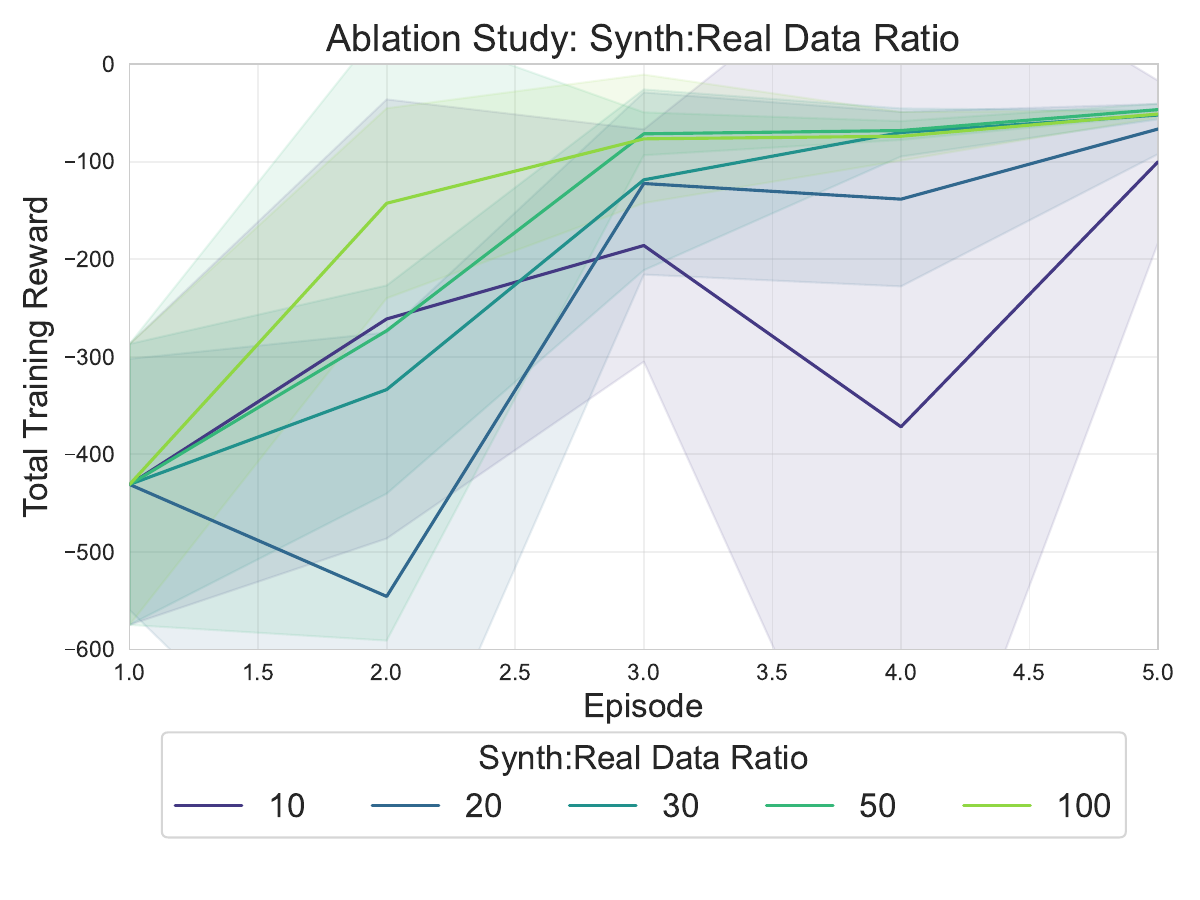}
    \caption{Learning curves for different synth:real data ratios.}
    \label{fig:study_synth_ratio}
\end{figure}

\begin{figure}
    \centering
    \includegraphics[width=0.48\textwidth]{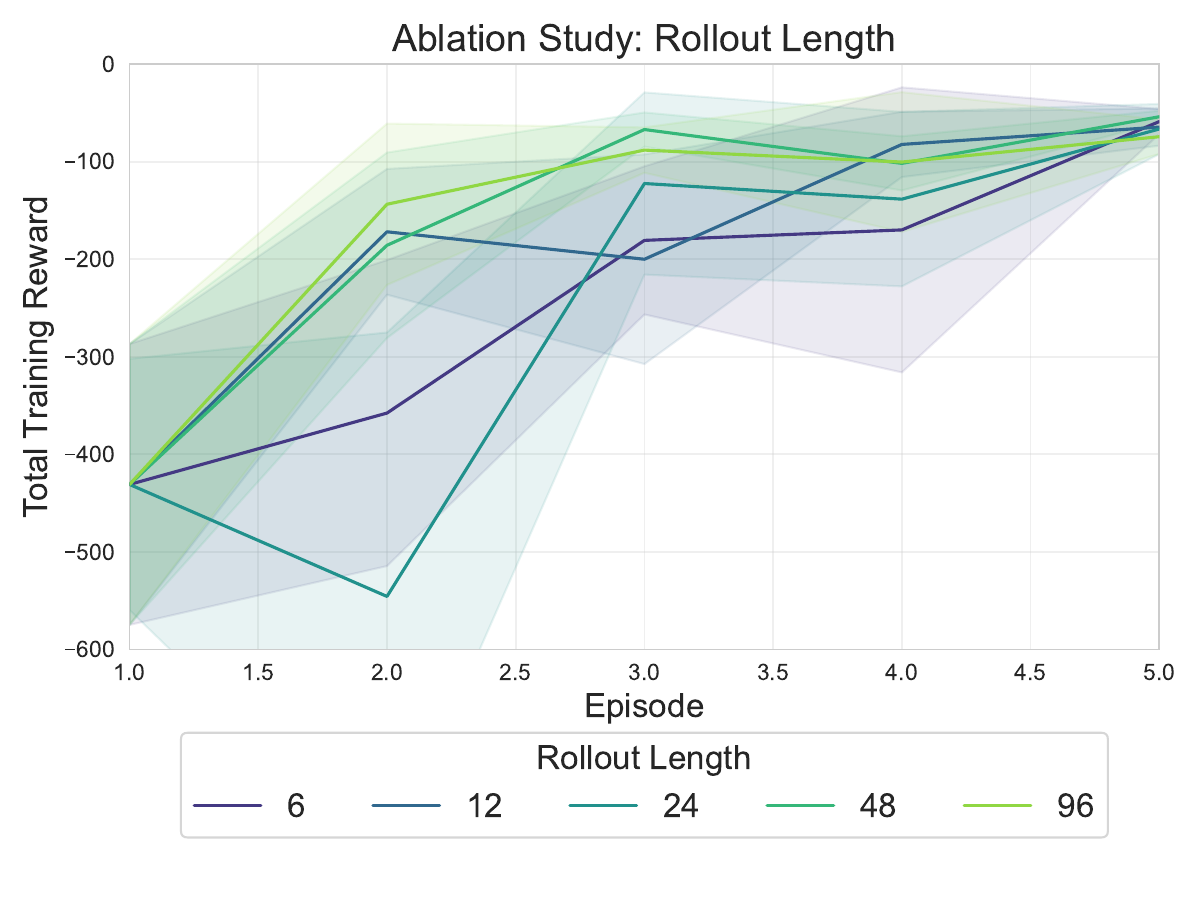}
    \caption{Learning curves for different rollout lengths.}
    \label{fig:study_rollout_len}
\end{figure}

\cref{fig:study_synth_ratio,fig:study_rollout_len} report learning curves in terms of average episodic reward. Increasing either the synth:real ratio or the rollout length tends to improve reward in the intermediate episodes, indicating faster learning from additional model-generated experience. However, after 5 episodes most configurations converge to a similar reward level; the main consistent outlier is the lowest synth:real setting (10), which underperforms relative to larger synthetic budgets.

\begin{figure}
    \centering
    \includegraphics[width=0.48\textwidth]{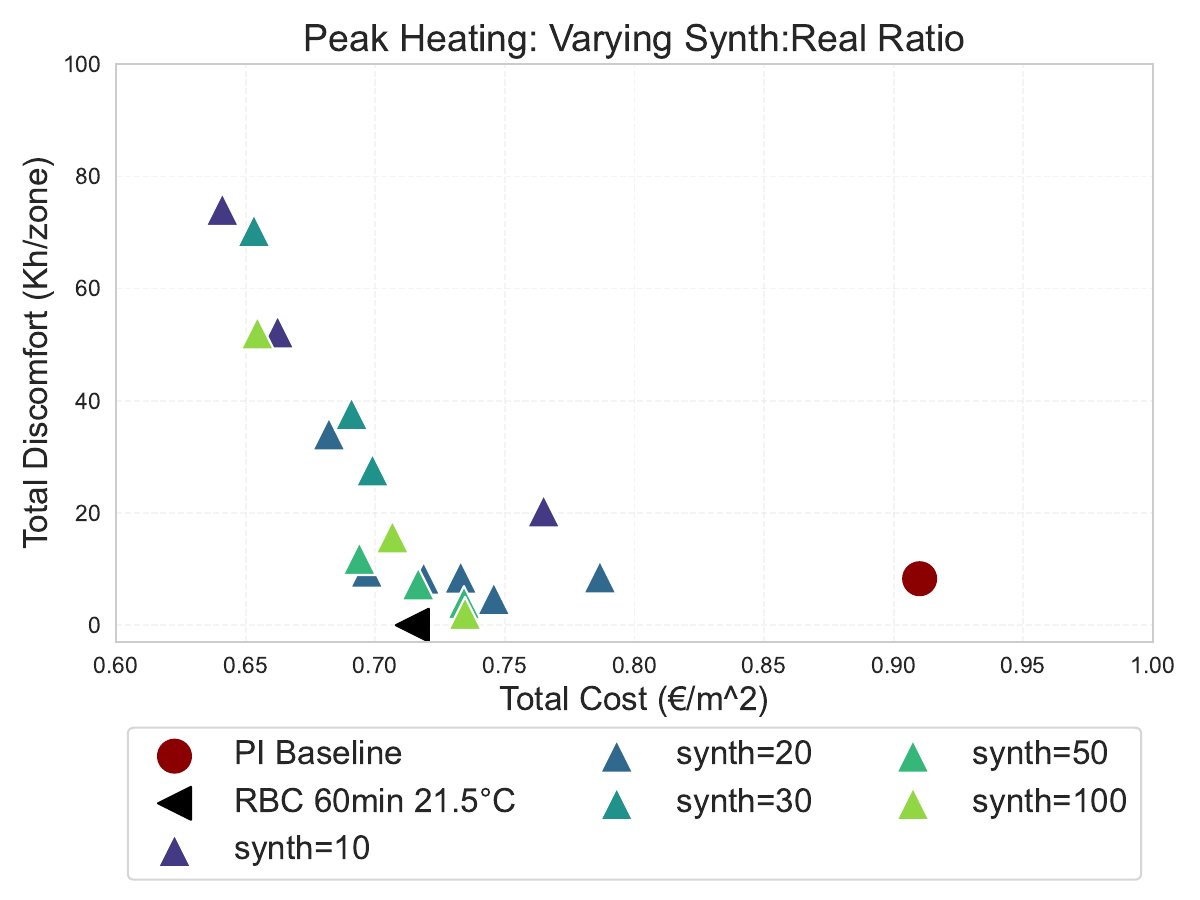}
    \includegraphics[width=0.48\textwidth]{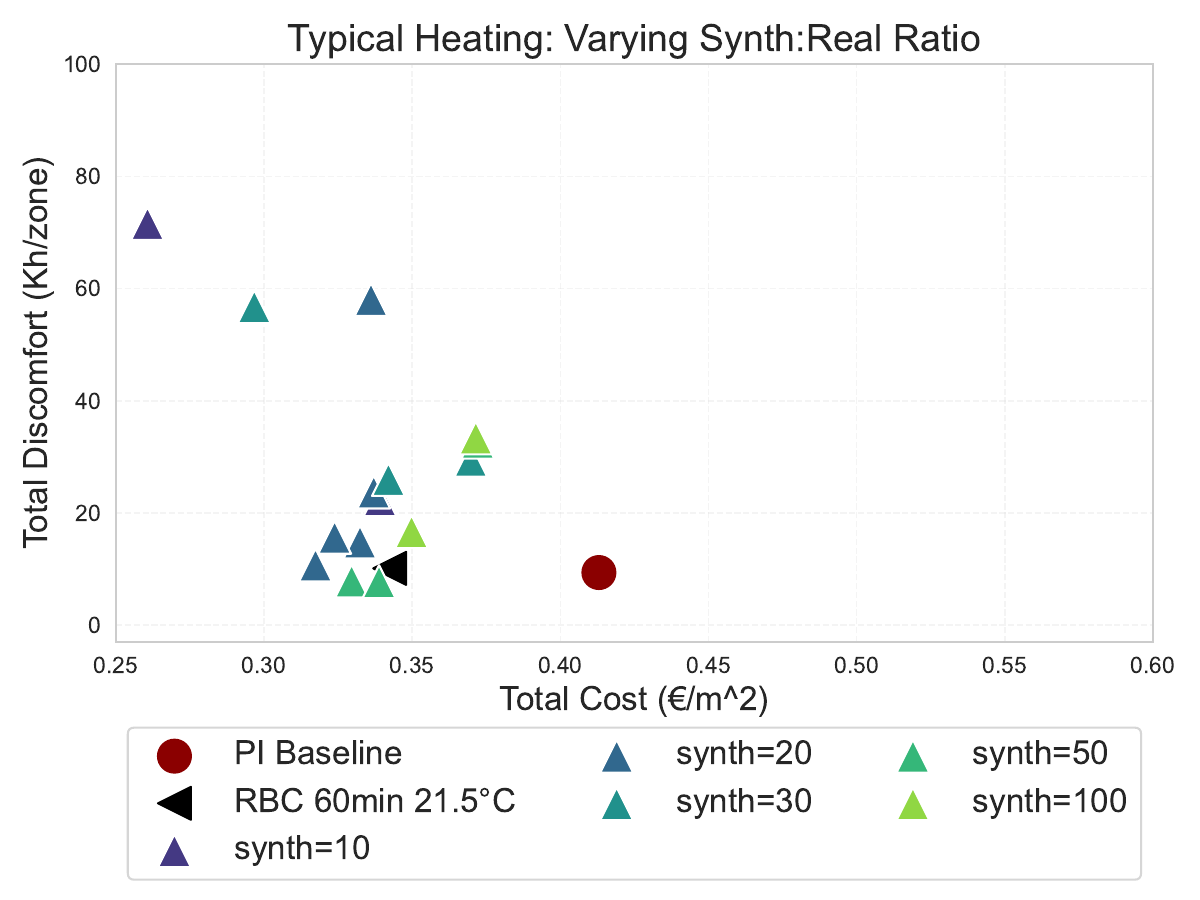}
    \caption{Cost vs. discomfort when varying the synth:real ratio (peak and typical heating tests).}
    \label{fig:ablation_scatter_synth_test_peak_heat_includingMF}
\end{figure}

\begin{figure}
    \centering
    \includegraphics[width=0.48\textwidth]{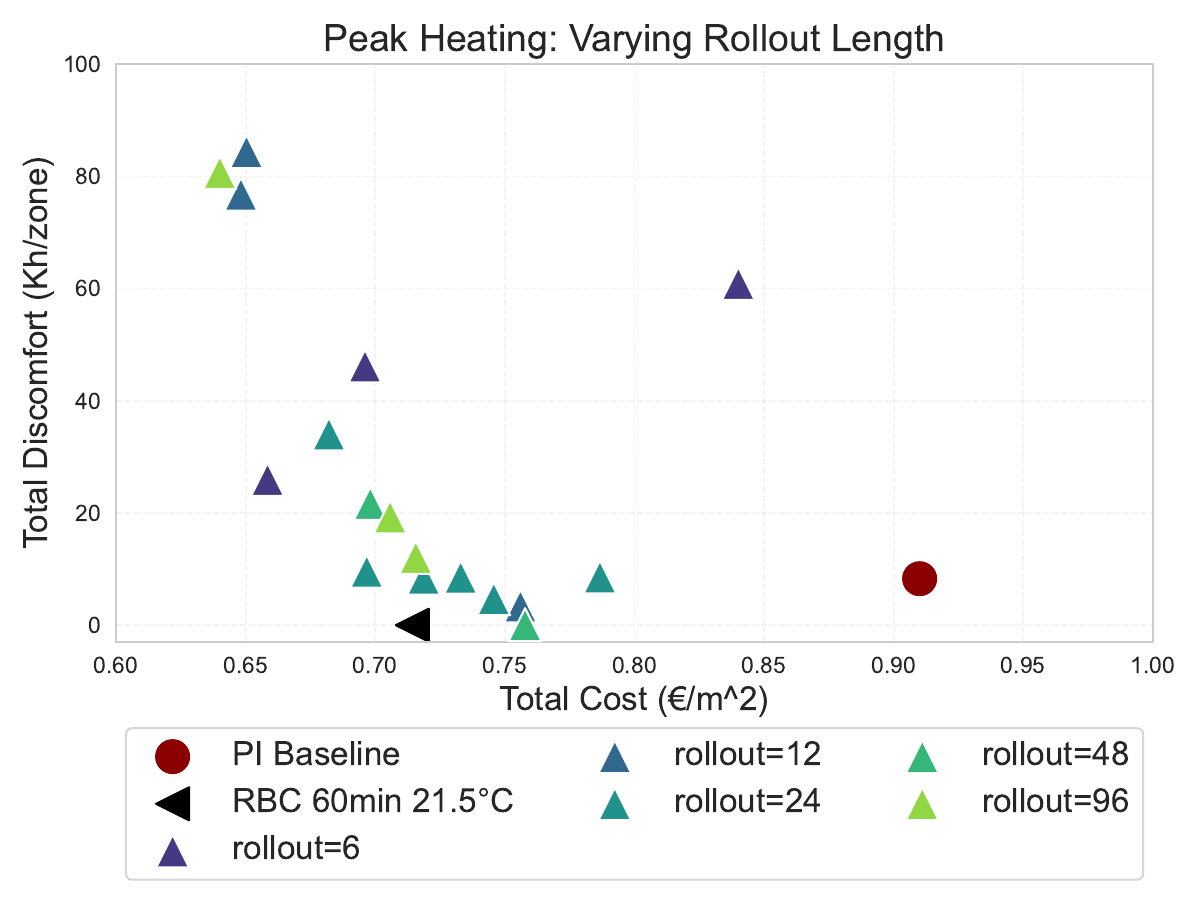}
    \includegraphics[width=0.48\textwidth]{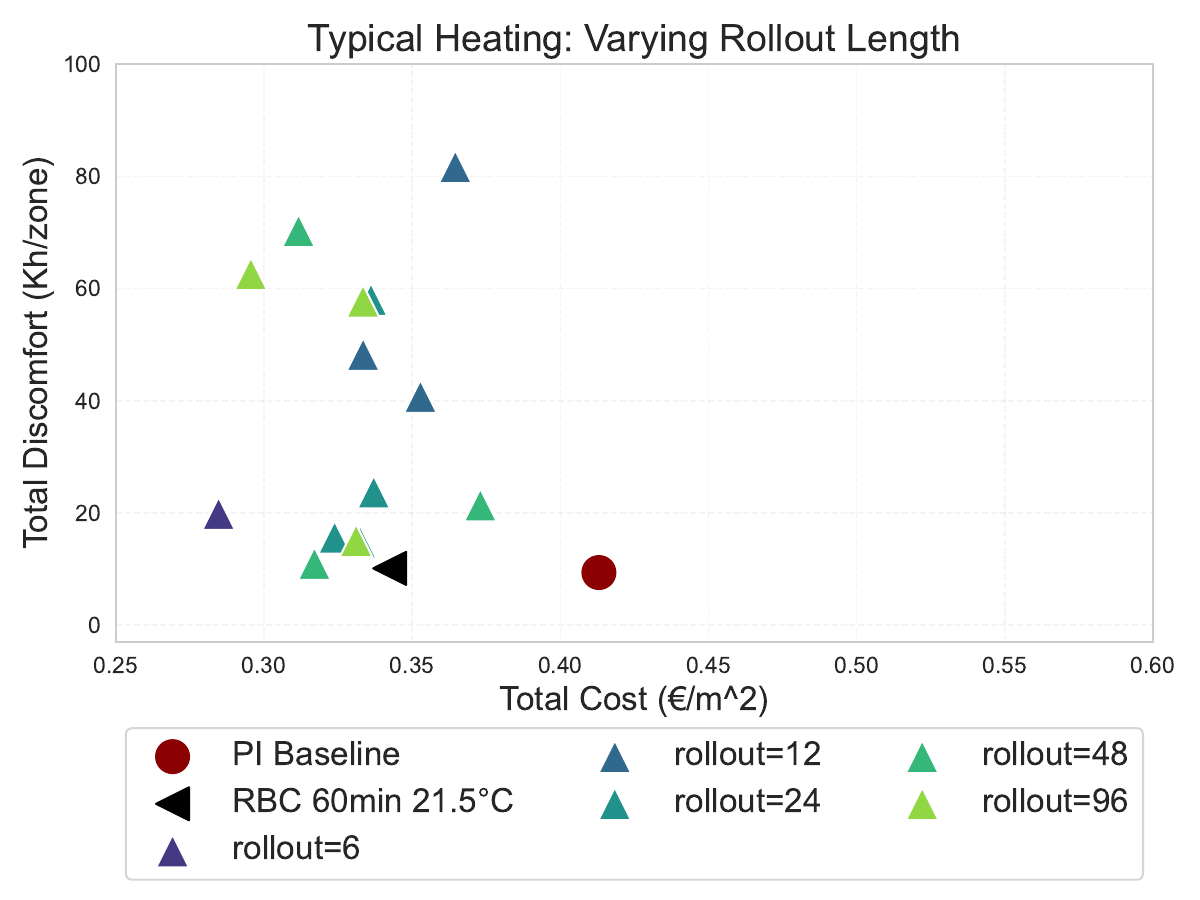}
    \caption{Cost vs. discomfort when varying rollout length (peak and typical heating tests).}
    \label{fig:ablation_scatter_rollout_test_peak_heat_includingMF}
\end{figure}

To assess whether these hyperparameters lead to meaningfully different operating points, \cref{fig:ablation_scatter_synth_test_peak_heat_includingMF,fig:ablation_scatter_rollout_test_peak_heat_includingMF} compare total cost and thermal discomfort across all runs for both test periods (peak heating and typical heating). Across seeds and configurations, the scatter plots show substantial overlap: no single synth:real ratio or rollout length yields a clearly dominant Pareto improvement in the cost--discomfort trade-off.

Overall, these results suggest that performance is relatively insensitive to the exact hyperparameter values within the studied ranges, provided that (i) the agent is trained with a sufficiently large amount of synthetic data (at least a 10:1 synth:real ratio in our experiments) and (ii) the rollout length is kept in a moderate regime, avoiding overly long rollouts that are more likely to accumulate model error.

\begin{figure}
    \centering
    \includegraphics[width=0.48\textwidth]{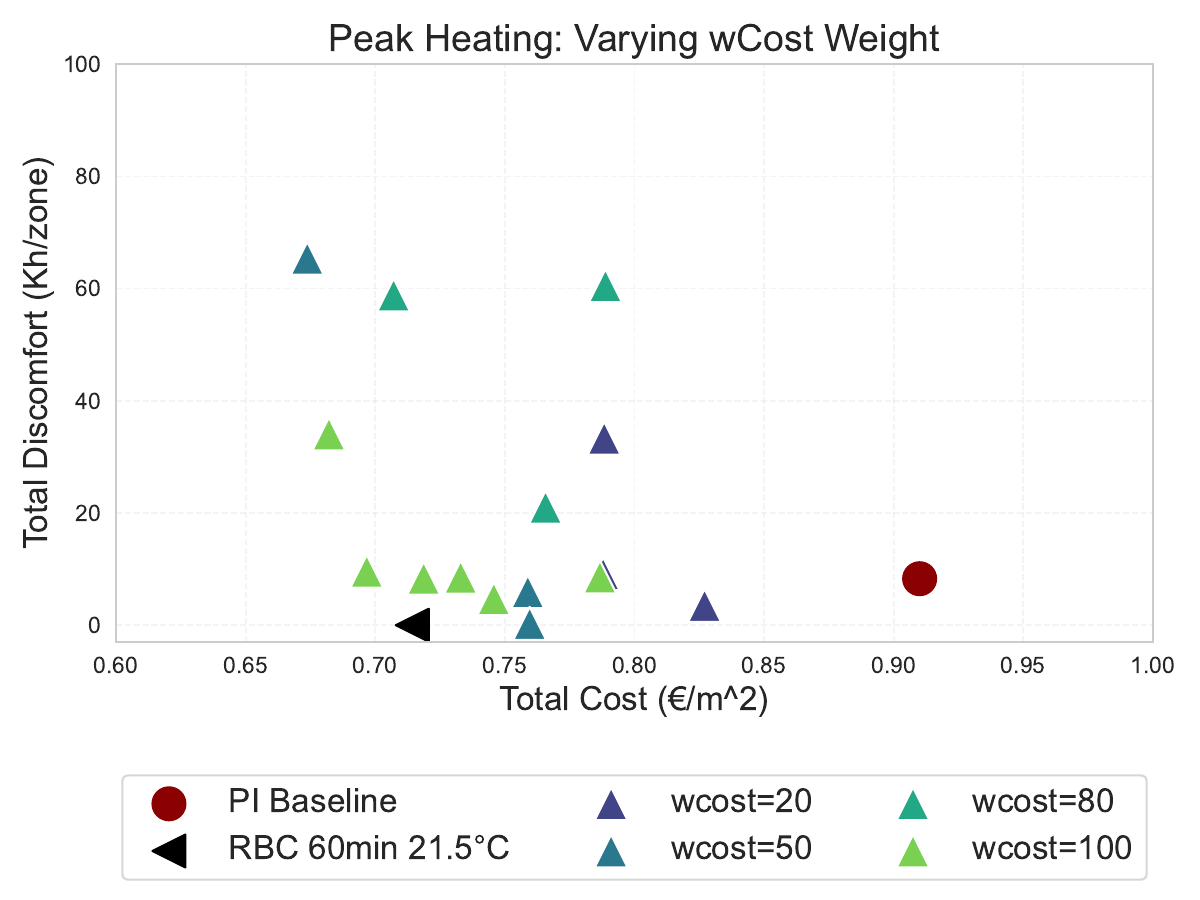}
    \caption{Cost vs. discomfort when varying the cost weight $w_C$.}
    \label{fig:scatter_wcost}
\end{figure}



\begin{table}[ht]
\caption{Synthetic:real data ratios and rollout lengths used in the ablation study. Other hyperparameters are kept the same as in \cref{tab:ppo_hyperparameters}. }
\label{tab:ablation_study_hparams}
\centering
\begin{tabular}{p{2.3cm} p{1.5cm}  p{4cm}}
\textbf{Hyperparameter} & \textbf{Value} & \textbf{Description} \\ \hline
Synth:real data ratio & [10, 20, 30, 50, 100] & Number of simulated steps per real step. (e.g. T = $20 \times K \times L$, see \cref{alg:counter_dyna})   \\ \hline
Rollout length ($L$) & [6, 12, 24, 48, 96] & Number of steps in a single synthetic episode. \\ \hline 
No. Episodes & 5 & Number of real episodes (1 episode = 1 week). \\
\end{tabular}
\end{table}

\clearpage
\newpage
\section{Hyperparameter Tables for Counter-Dyna and the CSM}
\label{sec:hyperparameter_tables}

\begin{table}[!ht]
\caption{Hyperparameters for the PPO RL agents.}
\label{tab:ppo_hyperparameters}
\centering
\begin{tabular}{p{2.3cm} p{1.5cm} p{4cm}}
\textbf{Hyperparameter} & \textbf{Value} & \textbf{Description} \\ \hline
Control step & 60min & Length of one environment step. \\ \hline
Actions & $\{0,1\}$ & Discrete activation signal sent to the heat pump. \\ \hline
Learning rate ($lr$) & 0.0005 & Constant step size for gradient updates. \\ \hline
Discount factor ($\gamma$) & 0.95 & Prioritizes immediate vs. long-term rewards. \\ \hline
Batch size & 21 & Size of the mini-batch for each gradient update. \\ \hline
$n_{steps}$ & 168 & Steps per environment rollout before updating. \\ \hline
$n_{epochs}$ & 10 & Number of times the optimizer passes through the rollout data. \\ \hline
Clip range & 0.3 & Limits the policy update to ensure stability. \\ \hline
Entropy coef. ($c_2$) & 0.01 & Entropy coefficient to encourage exploration. \\ \hline
Value function coef. & 0.25 & Weight of the value function (critic) loss. \\ \hline
Network Arch & $[128, 128, 128]$ & Three hidden layers for both policy and value networks. \\ \hline
Activation function & tanh & Activation function for the neural network layers. \\ \hline
Synth:real data ratio & 20 & Number of simulated steps per real step. (e.g. T = $20 \times K \times L$, see \cref{alg:counter_dyna})   \\ \hline
Rollout length ($L$) & 24 & Number of steps in a single synthetic episode. \\ \hline 
No. Episodes & $[5, 10]$ (Counter-Dyna), $[10, 50]$ (MF) & Number of real episodes (1 episode = 1 week). \\
\end{tabular}
\end{table}

\begin{table}[!ht]
\caption{Hyperparameters for the SAC RL agents.}
\label{tab:sac_hyperparameters}
\centering
\begin{tabular}{p{2.3cm} p{1.5cm} p{4cm}}
\textbf{Hyperparameter} & \textbf{Value} & \textbf{Description} \\ \hline
Control step & 60min & Length of one environment step. \\ \hline
Actions & [0, 1] & Continuous modulation signal for the heat pump. \\ \hline
Learning rate ($lr$) & 0.0005 & Constant step size for gradient updates. \\ \hline
Discount factor ($\gamma$) & 0.95 & Prioritizes immediate vs. long-term rewards. \\ \hline
Buffer size & 100,000 & Maximum number of experience samples stored. \\ \hline
Batch size & 64 & Size of the mini-batch for each gradient update. \\ \hline
Entropy coef. ($\alpha$) & auto & Automatically adjusted temperature parameter. \\ \hline
Tau ($\tau$) & 0.005 & Target network smoothing coefficient. \\ \hline
Train frequency & 1 & Number of steps between each update. \\ \hline
Gradient steps & 1 & Number of gradient updates per train step. \\ \hline
Learning starts & 100 & Steps taken before starting gradient updates. \\ \hline
Network Arch & [256, 256] & Hidden layer architecture for policy and value networks. \\ \hline
Activation function & tanh & Activation function for the neural network layers. \\ \hline
Synth:real data ratio & 20 & Number of simulated steps per real step. \\ \hline
Rollout length ($L$) & 24 & Number of steps in a single synthetic episode. \\ \hline 
No. Episodes & [5, 10] (Dyna) $\;\;$ [10, 50] (model-free) & Number of real episodes (1 episode = 1 week). \\
\end{tabular}
\end{table}

\begin{table}[ht]
\caption{Hyperparameters for the building model.}
\label{tab:bm_hyperparameters}
\centering
\begin{tabular}{p{2.1cm} p{1.9cm}  p{3.9cm}}
\textbf{Hyperparameter} & \textbf{Value} & \textbf{Description} \\ \hline
Learning rate ($lr$) & 0.001 & Constant step size for gradient updates. \\ \hline
Epochs ($lr$) & 500 & Number of times the agent does one pass through the training data. \\ \hline
Batch size & 256 & Size of the mini-batch for each gradient update. \\ \hline
Network Arch & [512, 512, 512] & Three hidden layers for both policy and value networks. \\ \hline
Activation function & LeakyReLU & Activation function for the policy and value network layers. \\
\end{tabular}
\end{table}


\end{document}